\documentclass{article}

    \PassOptionsToPackage{numbers, compress}{natbib}

    \usepackage[final]{neurips_2024}

\usepackage[utf8]{inputenc} 
\usepackage[T1]{fontenc}    
\usepackage{hyperref}       
\usepackage{url}            
\usepackage{booktabs}       
\usepackage{amsfonts}       
\usepackage{nicefrac}       
\usepackage{microtype}      
\usepackage{xcolor}         

\usepackage{graphicx}

\usepackage{multirow}
\usepackage{enumitem}
\usepackage{wrapfig}
\usepackage{caption}
\usepackage{algorithmic}
\usepackage{algorithm}

\title{CemiFace: Center-based Semi-hard Synthetic Face Generation for Face Recognition}

\author{%
\hspace{-1.2cm} Zhonglin Sun\\
\hspace{-1.2cm}  Queen Mary University of London\\
\hspace{-1.2cm} \texttt{zhonglin.sun@qmul.ac.uk} \\
  \And
  Siyang Song\thanks{Corresponding Author}  \\
  University of Exeter   \\
  \texttt{ss2796@cam.ac.uk} \\
  \AND
  Ioannis Patras\\
  Queen Mary University of London\\
  \texttt{i.patras@qmul.ac.uk} \\
  \And
  Georgios Tzimiropoulos\\
  Queen Mary University of London\\
  \texttt{ g.tzimiropoulos@qmul.ac.uk} \\
}

\begin{document}

\maketitle

\begin{abstract}

Privacy issue is a main concern in developing face recognition techniques. Although synthetic face images can partially mitigate potential legal risks while maintaining effective face recognition (FR) performance, FR models trained by face images synthesized by existing generative approaches frequently suffer from performance degradation problems due to the insufficient discriminative quality of these synthesized samples. In this paper, we systematically investigate what contributes to solid face recognition model training, and reveal that face images with certain degree of similarities to their identity centers show great effectiveness in the performance of trained FR models. Inspired by this, we propose a novel diffusion-based approach (namely \textbf{Ce}nter-based Se\textbf{mi}-hard Synthetic Face
Generation (\textbf{CemiFace})) which produces facial samples with various levels of similarity to the subject center, thus allowing to generate face datasets containing effective discriminative samples for training face recognition. Experimental results show that with a modest degree of similarity, training on the generated dataset can produce competitive performance compared to previous generation methods. The code will be available at:{\textcolor{red}{\href{https://github.com/szlbiubiubiu/CemiFace}{https://github.com/szlbiubiubiu/CemiFace}}}%

\end{abstract}

\section{Introduction}
\label{sec:intro}

Face Recognition (FR) has gained significant achievement in recent years owing to the combination of discriminative loss function~\cite{wang2018cosface,deng2019arcface,kim2022adaface,wen2021sphereface2,boutros2022elasticface}, proprietary backbones~\cite{Sun_2022_BMVC,xie2018comparator,li2023bionet,yang2020fan,wang2020hierarchical,sun2024lafs} and large-scale face datasets~\cite{zhu2021webface260m,cao2018vggface2,yi2014learning,guo2016ms}. For example, with a 4M training set, existing FR models can achieve over 99\% accuracy on various academic datasets~\cite{kim2022adaface,Sun_2022_BMVC,deng2019arcface,deng2021variational}. However, in real-world industrial face recognition applications, collecting large-scale face datasets is not always available due to the related licence agreements, ethical issues and privacy policies~\cite{regulation2016regulation}. 

To expand limited training samples in real-world scenarios, generative models~\cite{karras2019style,deng2020disentangled,esser2021taming,ho2020denoising,song2020denoising,rombach2022high} are widely adopted owing to their ability to generate high-quality images. However, simply adopting face images produced by those generic generative models to train face recognition models is impractical as there is ambiguity about the identities of the produced images because they are derived from random noises, i.e., the identities of these generated face images cannot be obtained without a well-trained FR model~\cite{kim2023dcface}. To address such issues, synthetic face dataset generation-based solutions~\cite{kim2023dcface,qiu2021synface,bae2023digiface,melzi2023gandiffface,boutros2023idiff} have been found to gain benefits in developing effective face recognition models. Existing synthetic face recognition (SFR) methods are frequently built upon recent advances of generative models such as Style-GAN~\cite{qiu2021synface,melzi2023gandiffface}, Diffusion methods~\cite{kim2023dcface,melzi2023gandiffface} and 3DMM rendering~\cite{bae2023digiface}. 
For instance, a style-transferring diffusion model-based method namely DCface~\cite{kim2023dcface} is proposed, which increases the diversity of existing face recognition datasets by generating additional discriminative face images with different styles (e.g. hair, overall lighting, which can be observed in visualization Section~\ref{subsec:qual_res}) for each subject. 
However, domain gap issues exists as the model is trained with paired face images belonging to the same identity, while those paired images are not available at the inference stage. It can only take samples belonging to different identities at the inference stage, which may negatively impact on the images synthesized at the inference stage. Furthermore, the definition of discriminative facial images remains unclear in this study.



\begin{wrapfigure}{r}{72mm}
  \centering
   \includegraphics[width=1.0\linewidth]{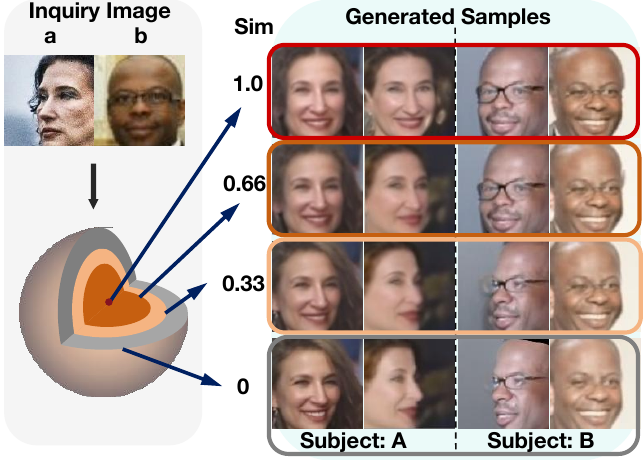}

    \caption{Visualization of the samples with different similarities. Given an inquiry image, it can form a hypersphere based on similarity to the inquiry image, where samples with the same similarity share the same radius. Samples with similarities between 0 to 1 with an interval of 0.33 are shown. With our proposed CemiFace, each inquiry image finally forms a novel subject. }
   
   \label{fig:diff_sim}
\end{wrapfigure}

To address the problems outlined above, firstly we explore the factors resulting in performance degradation for SFR and reveal that previous approaches fail to consider the properties of effective FR training--relationship/similarity between samples. Consequently, considering the facts: (a) semi-hard negative samples are crucial to train effective face recognition model for Triplet loss~\cite{schroff2015facenet}; (b) samples close to the decision boundary contribute most to the training gradient~\cite{kim2022adaface}; (c) all face images belonging to the same subject can be represented by a hypersphere in the latent feature space~\cite{liu2017sphereface} (i.e., can be measured by existing FR models, e.g., AdaFace~\cite{kim2022adaface}), whose distances (radius) to the identity center are negatively correlated to their similarities to the center. We hypothesize that face recognition performance is 
sensitive to the data with different levels of similarity to the identity center in the hypersphere, and experimentally reveal that the optimal performance is obtained with samples of mid-level similarity, which we term \textbf{center-based semi-hard samples}. Inspired by this crucial finding, we propose a novel diffusion-based synthetic face recognition approach (\textbf{CemiFace}) which generates \textbf{ce}nter-based se\textbf{mi}-hard face samples by regulating the similarity between the generated image and the inquiry image, through a similarity controlling factor condition. Figure~\ref{fig:diff_sim} presents the overall hypothesis by showcasing samples with various similarities to the identity center. Comprehensive experiments are conducted to illustrate the effectiveness of our proposed CemiFace. Our method achieves promising performance in synthetic face recognition (SFR). The main contributions and novelties of this work are summarized as follows:

\begin{itemize}

    \item We provide the first comprehensive analysis to illustrate how FR model performance is affected by different levels of similarity of samples, particularly center-based semi-hard samples.

    \item We propose a novel diffusion-based model CemiFace that can generate face images with various levels of similarity to the identity center, which can be further applied to generate infinite center-based semi-hard face images for SFR.
    
    
    \item We demonstrate our method can be extended to use as much as the data without label supervision for training which is an advantage over the previous method~\cite{kim2023dcface}.
    
    \item Experiments show that our CemiFace surpasses other SFR methods with a large margin, reducing the GAP-to-Real error by half.
    
    
    
    \end{itemize}





\section{Related Works and Preliminary}


\textbf{Synthetic Face Generation for FR:} 
With the emergency of generative models, synthesizing facial data for various facial tasks has become a critical issue, such as applications in Face Anti-sproofing~\cite{liu2020disentangling} and Face Recognition~\cite{qiu2021synface,bae2023digiface,kim2023dcface,boutros2023idiff,boutros2022sface,shen2018faceid}. SynFace~\cite{qiu2021synface} aims to mix the real images with the DiscoFaceGAN-generated~\cite{deng2020disentangled} samples. DigiFace~\cite{bae2023digiface} uses 3DMM for rendering facial images to construct the dataset. DCFace~\cite{kim2023dcface} takes diffusion models to adapt style from the style bank to the identity image and result in discriminative samples with diverse styles. IDiff-Face~\cite{boutros2023idiff} proposes the condition latent diffusion models~\cite{rombach2022high} to the feature embedding and images are synthesized by pretrained decoder.


\textbf{Preliminary-DDPM:} 
Diffusion models~\cite{ho2020denoising,song2020denoising} are generative models which denoise an image from a random noise image. The training pipeline for diffusion models consists of a forward process wherein noise is progressively added to a given image and a denoising process to predict the estimated noise for effective denoising. A single forward process is formulated as Markov Gaussian diffusion with timestep $t$:
\begin{equation}
    q(\mathbf{x}_{t}|\mathbf{x}_{t-1}) = \mathcal{N}(\mathbf{x}_{t};\sqrt{1-\beta_{t}} \mathbf{x}_{t-1},\beta_{t}\mathbf{I})
\end{equation}
Where $\mathcal{N()}$ is adding noise function. When accumulating the time step over $0-\mathbf{T}$, the final process is given as follows:
\begin{equation}
    q(\mathbf{x}_{1:\mathbf{T}}|\mathbf{x}_{0})=\prod^{\mathbf{T}}_{t=1}q(\mathbf{x}_{t}|\mathbf{x}_{t-1})
\end{equation}
Then the denoising process is conducted to predict the noise for the time step $t$ using a model $\mathbf{\sigma_{\theta}}$~( typically a UNet~\cite{ronneberger2015u}), the training loss is:
\begin{equation}
    L_{\mathbf{MSE}}=E_{t,\mathbf{x_{0}},\epsilon}[||\sigma_{\theta}(\sqrt{\bar{\alpha}_{t}}\mathbf{x_{0}}+ \sqrt{1-\bar{\alpha}_{t}}\epsilon,t)-\epsilon||^{2}_{2}]
\end{equation}\label{equation:mse}
where $\beta_{t}$ is the pre-set forward process variances. Then notation $\bar{\alpha}_{t}$ is given as: $\bar{\alpha}_{t} = \prod^{t}_{s=1} \alpha_{s}$ and $\alpha_{t}=1- \beta_{t}$. $\epsilon$ is a random Gaussian noise image $\mathbf{\epsilon}\sim \mathcal{N}(0,1)$.

\section{The proposed approach}
\label{sec:method}

In Section~\ref{Sec:method_semi_hard}, we first investigate the relationship between sample similarity and their effectiveness in training FR models, presenting the finding that samples with certain similarities (i.e., center-based semi-hard samples) to their identity centers are more effective for training FR models on a real dataset and subsequently devise a toy experiment to validate it. Inspired by our findings,  in Section~\ref{subsec:ourmethod} we propose a novel CemiFace, a conditional diffusion model that produces images with various levels of similarity to an inquiry image. Specifically, Section~\ref{Sec:train_cemi} introduces how we construct the similarity condition which is fed to diffusion model to guide the generation, and discusses the $L_{SimMat}$ to require the generated sample to exhibit a certain similarity degree to the inquiry image. In Section~\ref{Sec:gen_sim}, we then present how to use our diffusion model to generate a synthetic face dataset given a fixed similarity condition $m$ and a set of inquiry images.

\subsection{The Relationship between Samples Similarity and Performance Degradation}
\label{Sec:method_semi_hard}

\textbf{Performance Degradation for Synthetic Face Recognition:} Face recognition models trained on face images synthesised by existing generative models (e.g., style-transfering~\cite{kim2023dcface}, 3DMM rendering~\cite{bae2023digiface} and latent diffusion expansion~\cite{boutros2023idiff}) frequently suffer from performance degradation~\cite{kim2023dcface,qiu2021synface,bae2023digiface}. For example, with the same data volume, the model trained on the state-of-the-art synthetic dataset DCface~\cite{kim2023dcface} produces 11.23\% lower verification performance on CFP-FP testset than the model with the same architecture trained on the real dataset. A key reason for this issue is that these generative models only intuitively explore the properties of discriminative samples, but fail to consider the similarity levels among synthesized face images. However, previous studies~\cite{schroff2015facenet,deng2019arcface,kim2022adaface} empirically reveal that \textit{training effective FR models intrinsically relies on semi-hard negative samples in Triplet Loss~\cite{schroff2015facenet} or samples close to the decision boundary~\cite{deng2019arcface,kim2022adaface,meng2021magface}}.

\begin{table}[h]
\begin{minipage}[b]{0.45\linewidth}
\includegraphics[width=1.0\columnwidth]{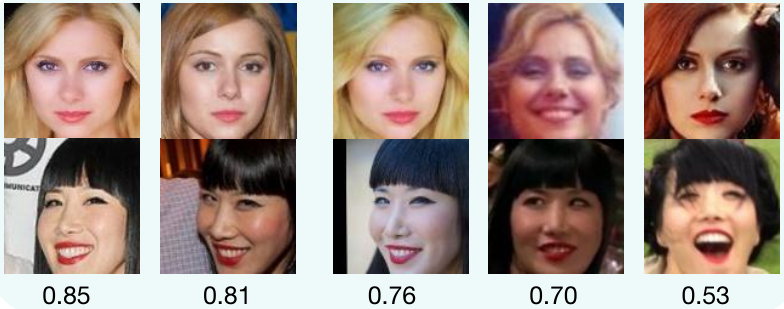}
\captionof{figure}{Samples with different similarity groups from CASIA-WebFace dataset. From left to right are samples with lower similarity to the identity center}
\label{fig:CASIA_sim_group_img}
\end{minipage}
\hspace{2mm}
\begin{minipage}[b]{0.5\linewidth}

\resizebox{1.0\columnwidth}{!}{
\begin{tabular}{@{}l|lllll|l@{}}
\toprule
Sim & LFW            & CFP-FP         & AgeDB      & CALFW         & CPLFW          & AVG             \\ \midrule
0.85  & 98.43          & 85.67          & 89.43         & 91.08         & 82.78          & 89.48          \\
0.81  & 98.91          & 88.8           & 91.03         & 91.71         & 84.58          & 91.01          \\
0.76  & \textbf{98.94} & 90.92          & \textbf{91.5} & \textbf{91.7} & 85.85          & \textbf{91.78} \\
0.70  & 98.66          & \textbf{91.08} & 90.32         & 90.76         & \textbf{86.92} & 91.55          \\
0.53  & 94.63          & 82.12          & 77.63         & 80.11         & 77.3           & 82.36          \\ \bottomrule
\end{tabular}}
\caption{Accuracy of groups with different similarities. Sim means the average similarity to the identity center. AVG is the average accuracy on the 5 evaluation datasets}
\label{Tab:casia_semihard}
\end{minipage}  

\end{table}

\textbf{Hypothesis and Findings:} \label{Sec:hypo_finding} Since face images belonging to the same identity/class can be aggregated within a hypersphere~\cite{liu2017sphereface}, where the location of each face image is decided by its similarity to the identity center (the center of the hypersphere) (illustrated in Fig. \ref{fig:diff_sim}). We treat all face images of each subject as an N-1 (N=512 in AdaFace~\cite{kim2022adaface}) dimensional sphere with its center representing the subject-level identity center. Then, the spheres of all subjects can be combined in an N-dimensional sphere, where each subject-level sphere is a cluster. 

Based on this, we hypothesize that samples of mid-level similarities to the identities center play a dominant impact on the FR performance, as they exhibit discriminative style variations (e.g. age, pose). To validate the hypothesis, we conduct the first comprehensive investigation for the impact of different levels of similarity to the identity center on the FR performance. We first split face images in the CASIA-WebFace~\cite{li2017learning} into various levels of groups according to their similarities to their corresponding subject-level identity centers. Here, the identity center of each subject is obtained by the weight of the linear classification layer, trained using AdaFace~\cite{kim2022adaface}.
To avoid the impacts caused by different numbers of training samples, we assign around 100k face images to each group representing close similarities to their identity centers. This results in 5 distinct similarity groups. Table~\ref{Tab:casia_semihard} reports the performance of model trained on each group and test on five standard face recognition evaluation datasets \cite{huang2008labeledlfw,sengupta2016frontalCFP,moschoglou2017agedb,zheng2018cross,zheng2017cross}. Table~\ref{Tab:similarity_range} in \textit{Supplementary material} Section~\ref{Sec:sim_range} displays the similarity range of each group. We further validate the style variation in Visualization Sec~\ref{subsec:qual_res}.
We also visualize randomly selected samples of each group in Figure~\ref{fig:CASIA_sim_group_img}. Results reveal that groups with middle-level similarities (0.76 and 0.70) produced similar but top-performing average accuracy. This indicates that \textbf{face images of a certain low similarity to their identity centers (which we refer to as center-based semi-hard samples) are essential for learning highly accurate face recognition models.} In contrast, the group whose images have the lowest similarity (i.e., 0.53) to their identity centers obtained the worst performance, which suggests that it is difficult to train an effective face recognition model with the most challenging samples (i.e., the samples are normally hard to be distinguished by human observation).

\subsection{Center-based Semi-hard Face Image Generator}
\label{subsec:ourmethod}

Inspired by the above findings, this section proposes a novel conditional diffusion model, namely CemiFace, for synthesising effective center-based semi-hard face images given the inquiry image (identity center) $\mathbf{x}$ and a pre-defined similarity controlling factor $\mathbf{m}$, based on which a new discriminative synthetic dataset is obtained to train effective face recognition models. 


\begin{figure*}[htbp]
  \centering
   \includegraphics[width=0.98\linewidth]{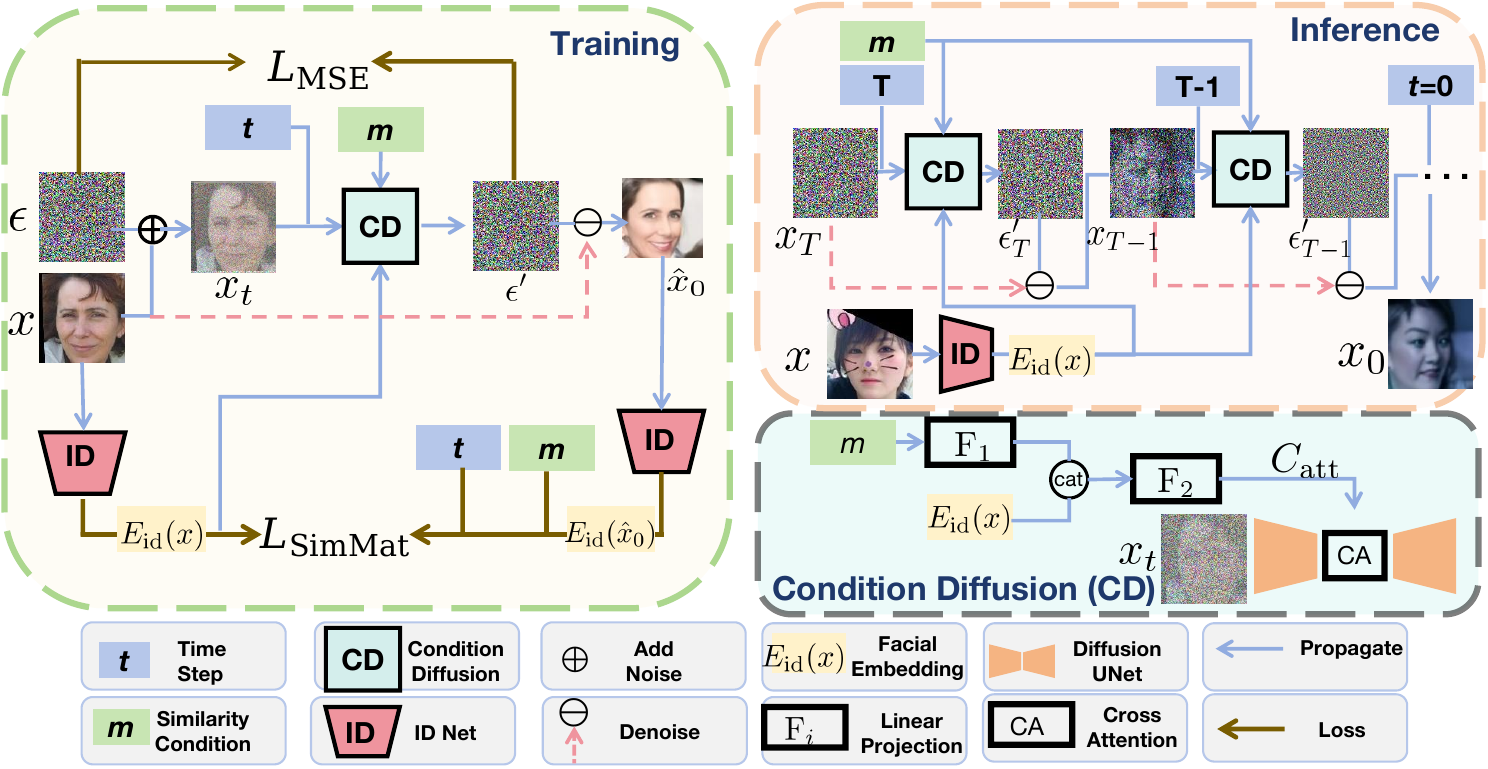}

    \caption{Illustration of our proposed method. The left part is the training framework for learning images with various levels of similarity. Firstly noise is added to the clean facial image before it is processed by the diffusion model. Then similarity controlling condition $\mathbf{m}$ ranging between [-1,1] with facial embedding is injected to guide the generation. Consequently, the model outputs the estimated noise, which is adopted to calculate the estimated image. We add similarity matching loss $L_{\mathbf{SimMat}}$ between the estimated image and the input image. For generation, we gradually denoise a noising image with time step scaling from $\mathbf{T}$ to 0, conditions for identity and similarity are left fixed. The two diffusion models in the generation part mean the same diffusion model at two different time steps. The right bottom part is the details of using cross-attention to inject similarity condition and facial embedding into the diffusion models}
   
   \label{fig:method_af_rebu}
\end{figure*}
\paragraph{Methodology overview:} As illustrated in the left side of Fig~\ref{fig:method_af_rebu}, the training process starts with adding a noise $\epsilon \sim \mathcal{N}(0,1)$ with timestep $t$ to the clean input image $x$, resulting in $x_{t}$. Meanwhile, similarity conditions $m$ and identity condition $\mathbf{E_{id}}(x)$ are fed to the diffusion model by cross-attention as illustrated in the lower right part of Fig.~\ref{fig:method_af_rebu}. 
Consequently, the diffusion Unet $\sigma_{\theta}$ outputs the estimated noise $\epsilon'=\sigma_{\theta}(x_{t},t,m, \mathbf{E_{id}}(x))$ for denoising the image as a clean estimated image $\hat{x}_{0}$. Based on the obtained estimated image $\hat{x}_{0}$, original $x$ and condition $C_{att}$, the whole model is optimized by the combination of $L_{MSE}$ and $L_{SimMat}$, defined in the following Section~\ref{Sec:train_cemi} as well as the details of constructing condition.


At the inference stage (the upper right part of Fig.~\ref{fig:method_af_rebu}), random noise $x_{t}=x_{\mathbf{T}}=\epsilon \sim \mathcal{N}(0,1)$ and the time step $t=\mathbf{T}$ are first fed to a CD block. This results in an estimated noise $\epsilon'=\sigma_{\theta}(x_{t}, t, \mathbf{C_{att}})$. Then, a denoise step is adopted to generate $x_{t-1}$ from $x_{t}$ for efficient interface speed. This process is repeatedly conducted on the obtained denoised latent images ($x_{t-1}, x_{t-2}, \cdots, x_{0}$) until $t=0$, where $x_{0}$ is treated as the final generated face image.
Here, we assign the same identity label as $x$ to all face images generated from the inquiry image $x$. To ensure high inter-class variation, our inquiry images are filtered by a pretrained FR ( IR-101 trained on the WebFace4M~\cite{zhu2021webface260m} dataset by AdaFace.), which enforces the similarity between each pair of query images is lower than 0.3. The number of identities is fully decided by the number of inquiry face images. The pseudo-code for training and generation are given in \textit{Supplementary Material} Section~\ref{sec:pseudo}.

\subsubsection{Training CemiFace}\label{Sec:train_cemi} 

To facilitate our diffusion-based CemiFace can generate diverse center-based semi-hard face images, we propose a novel diffusion model training strategy. During training, a random Gaussian noise image $\mathbf{\epsilon}\sim \mathcal{N}(0,1)$ is firstly added to a clean face image $\mathbf{x}$ at the time step $t$, before feeding it to the diffusion model to generate the noise face image $\mathbf{x_{t}}$:
\begin{equation}\label{eq:add_nosie}
    \mathbf{x_{t}}=\sqrt{\mathbf{\bar \alpha_{t}}}\mathbf{x_{0}}+\sqrt{1-\bar \mathbf{\alpha_{t}} }\mathbf{\epsilon}
\end{equation}
Then, conditions are constructed based on the similarity controlling factor $\mathbf{m}$, the identity condition $\mathbf{C}_{\mathbf{id}}$ and time step $t$ condition. Subsequently, the diffusion model outputs the estimated noise $\mathbf{\epsilon'}=\mathbf{\sigma_{\theta}}(\mathbf{x}_{t},t,\mathbf{E_{id},\mathbf{m}})$ for denoising the image.
    




\textbf{Constructing Similarity Controlling Condition:}
\label{Sec:sim_cond}
To address the purposes of generating images at different scales of similarities, two conditions are injected into the diffusion process to guide the generation process. The first one is the identity condition $\mathbf{C}_{\mathbf{id}}$ aiming to anchor the center of the generated facial images which can be formulated as: 
\begin{equation}
    \mathbf{C}_{\mathbf{id}}= E_{\mathbf{id}}(\mathbf{x})
\end{equation}
where $E_{\mathbf{id}}$ is a pre-trained face recognition model (e.g., IResnet-50 pretrained from AdaFace~\cite{kim2022adaface}). $\mathbf{C_{\mathbf{id}}}$ represents the feature embedding of the given image $\mathbf{x}$. Then the most important part is similarity controlling condition $\mathbf{C_{sim}}$ which maps the scalar similarity $\mathbf{m}$ into feature embedding. This condition serves to regulate the similarity to the inquiry image, facilitating the generation of images spanning from the most challenging samples ($\mathbf{m}$=-1) to the most similar ones ($\mathbf{m}$=1).

\begin{equation}
    \mathbf{C_{sim}}=F_{1}(\mathbf{m)}
\end{equation}
Where $F_{i}()$ is the linear projection layer. Then following DCFace~\cite{kim2023dcface} the two conditions are combined and projected as cross-attention conditions for sending to the DDPM process. AdaGN~\cite{preechakul2022diffusion} is adopted to embed time step condition $t$. $\mathrm{cat()}$ is the concatenation operation.
\begin{equation}\label{eq:attention}
    \mathbf{C_{att}}=F_{2}(\mathrm{cat}(\mathbf{C_{id}},\mathbf{C_{sim}}))
\end{equation}

The $\mathbf{\mathbf{C_{att}}}$ is further processed by a cross-attention operation with the intermediate latent representation of diffusion UNet $\sigma_{\theta}$ learned from the input noisy image as: 
\begin{equation}
    CA(Q,K,V,K_{c},V_{c})=SoftMax(\frac{QW_{q}([K,K_{c}]W_{k})^{T}}{\sqrt{d}})W_{v}[V,V_{c}]
\end{equation}
where $\mathbf{C_{att}}$ is treated as the key $K_{c}$ and value $V_{c}$ (same as DCFace) to influence the generated face images. $Q=K=V$ are the query, key and value, representing the latent feature of UNet $\sigma_{\theta}$.

\textbf{Training Loss:} To ensure the similarity between the generated face $\mathbf{x_{0}}$ and the corresponding inquiry image (identity center) $\mathbf{x}$ adheres to the specified similarity factor $\mathbf{m}$ as given in the following equation:
\begin{equation}
    \mathbf{m} = \mathrm{sim}(E_{\mathbf{id}}(\mathbf{x}),E_{\mathbf{id}}(\mathbf{x}_{\mathbf{0}}))
\end{equation}

where $\mathrm{sim()}$ denotes a similarity measurement (e.g., can be computed by Cosine Similarity or Euclidean Distance). Following DDPM~\cite{kim2023dcface,ho2020denoising}, an approximated clean sample $\mathbf{x_{0}}$ can be traced from $\mathbf{x}_{t}$ at the time step $t$ through the following formula:
\begin{equation}\label{eq:estimated_x0}
    \mathbf{x_{0}}\approx \mathbf{\hat{x}_{0}}=(\mathbf{x}_{t}-\sqrt{1-\bar{\mathbf{\alpha}_{t}}}\mathbf{\epsilon'})/\sqrt{\bar{\mathbf{\alpha}_{t}}}
\end{equation}

This gives a hint that the generated face image $\mathbf{x_{0}}$ can be controlled at the training phase by regularizing the estimated $\mathbf{\hat{x}_{0}}$, which allows the gradient to be back-propagated to the diffusion model, e.g., controlling facial attributes~\cite{zeng2023face} and styles~\cite{kim2023dcface}. Inspired by this, we propose a novel similarity Matching loss $L_{\mathbf{SimMat}}$ aimed at disentangling the generated face image $\mathbf{x_{0}}$ to exhibit a certain similarity to the inquiry image, which is determined by the similarity controlling factor $\mathbf{m}$. We employ the Time-step Dependent loss~\cite{kim2023dcface} with different time step t at Eq~\ref{eq:simmat}, specifically firstly an identity loss for recovering the identity of the original inquiry image x, which will be applied to produce original facial embedding when the time step $t\rightarrow 0$:

\begin{equation}
    L_{\mathbf{rec}}=||1-\mathrm{sim}(E_{\mathbf{id}}(\mathbf{x}),E_{\mathbf{id}}(\hat{\mathbf{x}}_{0}))||_{2}
\end{equation}
Then, we require the estimated $\hat{\mathbf{x}}_{0}$ to produce an feature embedding $E_{\mathbf{id}}(\hat{\mathbf{x}}_{0})$ which matches the original $\mathbf{x}$ with $\mathbf{m}$ similarity as:
\begin{equation}
    L_{\mathbf{sim}}=||\mathbf{m}-\mathrm{sim}(E_{\mathbf{id}}(\mathbf{x}),E_{\mathbf{id}}(\hat{\mathbf{x}}_{0})||_{2}
\end{equation}
Consequently, the overall identity regularization loss at the time step $t$ can be formulated as:
\begin{equation}\label{eq:simmat}
    L_{\mathbf{SimMat}}=(1-\gamma_{t}) L_{\mathbf{rec}} + \gamma_{t} L_{\mathbf{sim}}
\end{equation}
where $\gamma_{t}=\frac{t}{\mathbf{T}}$ is the scaling weight for adjusting the similarity of the generated $\mathbf{\hat{x}}_{0}$. At the time step $t$=0, the model outputs an image with the same identity as the original image $\mathbf{x}$. When $t$ scales from 0 to the maximum time step $\mathbf{T}$, the generated face image gradually shifts far away from the $\mathbf{x}$. When approaching $\mathbf{T}$, the model will output the image with $\mathbf{m}$ similarity to the original image.  The proposed $L_{\mathbf{SimMat}}$ loss is inspired by the fact that facial images, with diverse styles but the same degree of similarity, are located at a circle of the hypersphere. This loss can regularize the model to learn this kind of pattern. Specifically, the similarity is guaranteed by our proposed loss, and the diversity is facilitated by the random noise $\mathbf{\epsilon}$ of the diffusion models, which is validated in the Visualization Sec.~\ref{subsec:qual_res}. The overall training object is:
\begin{equation}\label{eq:total_loss}
    L=L_{\mathbf{MSE}}+ \lambda L_{\mathbf{SimMat}}
\end{equation}
where $\lambda$ is a hyperparameter for balance the training focuses on noise estimation or identity-related similarity regularization.

\subsubsection{Face Image Generation with Appropriate Similarity}\label{Sec:gen_sim}

Given a random noise $\mathbf{\epsilon}$ and conditions(i.e. identity, similarity and time step), the well-trained model progressively denoises the noisy image $\mathbf{\epsilon}$ with a varying time step $t$ (from the maximum $\mathbf{T}$ to 0) and a fixed similarity factor condition $\mathbf{m}$ to generate a clean image with a specified similarity $\mathbf{m}$ to the given inquiry image $\mathbf{x}$, we adopt DDIM~\cite{song2020denoising} for efficient interface speed.


We experimentally investigate the appropriate generation similarity $\mathbf{m}$ for synthetic face recognition. Specifically, we first adopt fixed similarity factors to test the best similarity. We also explore mixing the similarity around the appropriate fixed $\mathbf{m}$ (mixing semi-hard $\mathbf{m}$) and mixing appropriate fixed $\mathbf{m}$ samples with easy samples (mixing easy $\mathbf{m}$).

\section{Experiment}

\subsection{Implementation Details}




\textbf{Evaluation Metrics:}  We examine the 1:1 verification accuracy trained on the dataset generated by our CemiFace on various famous testsets including LFW~\cite{huang2008labeledlfw}, CFP-FP~\cite{sengupta2016frontalCFP}, AgeDB-30~\cite{moschoglou2017agedb}, CPLFW~\cite{zheng2018cross}, CALFW~\cite{zheng2017cross} and their average verification accuracy $\mathbf{AVG}$. Gap-to-Real is the gap to the results trained on CASIA-WebFace with CosFace loss.

\textbf{Details of CemiFace Training and Generation:} The condition $\mathbf{m}$ is appropriately adjusted during the training phase to facilitate better generalization across various similarities. Considering the overall cosine similarity ranges from -1 to 1, the model is enabled to discern differences in generated images under varying similarity controlling conditions when training. Specifically, in the mini-batch, we assign a randomly selected m from -1 to 1 with an interval of 0.02, allowing the model to generate corresponding images at different similarity scales. The synthetic face recognition datasets are generated in 3 volumes. Specifically in 0.5M data volume, we generate 50 images per subject and a total of 10k subjects; As for 1.0M, we keep 50 images per subject but with 20k subjects; For 1.2M, we add 5 images per subject with 40k subjects to the 1.0M settings. Oversampling method as used in DCFace is adopted which adds 5 repeated inquiry images to each subject. For more details including model, ablation studies and discussions please refer to \textit{Supplementary material}~\ref{Sec:runtime},~\ref{Sec:further_dis} and~\ref{Sec:pri_con}




\begin{wrapfigure}{r}{75mm}
\centering
\includegraphics[width=1.0\linewidth]{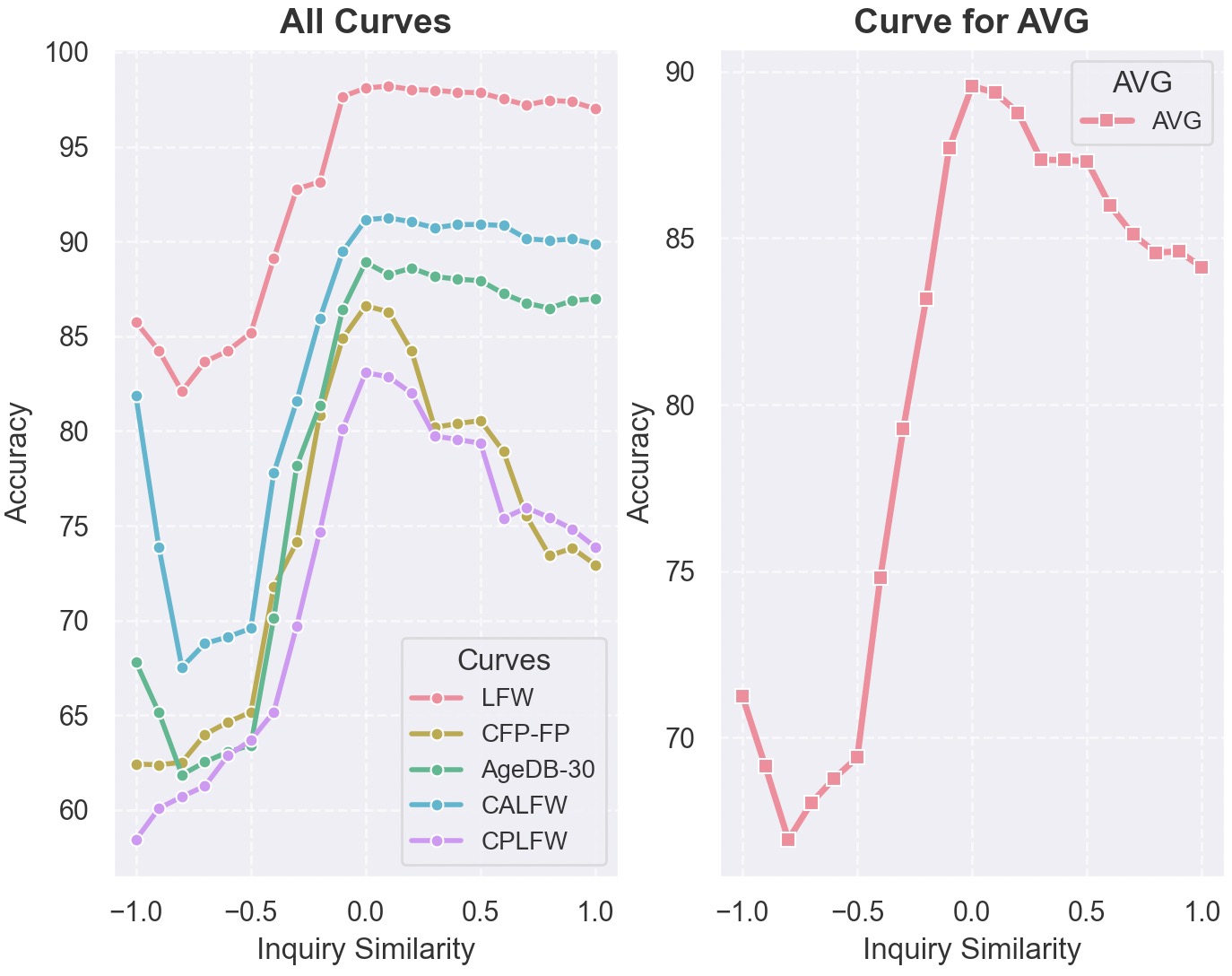}
\caption{Accuracy of samples with different similarity varying from -1 to 1. The left figure is the specific performance on each evaluation dataset. The right figure is the average accuracy of our CemiFace}
\label{fig:CASIA_sim_group}
\vspace{-14mm}
\end{wrapfigure}

\textbf{Details of Training the Synthetic Dataset}\label{Sec:train_details}
As the training code of DCFace~\cite{kim2023dcface} and DigiFace~\cite{bae2023digiface} for training the SFR is not released. We opt for CosFace~\cite{wang2018cosface} with some regularizations to match the performance of DCFace~\cite{kim2023dcface}. Specifically, the margin of Cosface is 0.4, weight decay is 5e-4, learning rate is 1e-1 and is decayed by 10 at the 26th and 34th epoch, totally the model is trained for 40 epochs.  We add random resize~\& crop with the scale of [0.9, 1.0], Random Erasing with the scale of [0.02,0.1], and random flip. Brightness, contrast, saturation and hue are all set to be 0.1. The backbone opted for is IR-SE50~\cite{deng2019arcface}

\subsection{Ablation Studies}\label{sec:aba}
\subsubsection{Impact of Similarity m}\label{aba:inq_img}

\textbf{Appropriate m for Generation:}\label{ab:m} Herein we ablate how a scalar $\mathbf{m}$ influences the generation in terms of training performance. We adopt different $\mathbf{m}$ ranging from -1 to 1 with the interval of 0.1 to generate face groups of 10k identities with 10 samples per identity to match the data volume in the finding for CASIA-WebFace in Sec.~\ref{Sec:method_semi_hard}. Figure~\ref{fig:CASIA_sim_group} illustrates the accuracy curves when using those data for training face recognition (for detailed numerical results please refer to \textit{Supplementary Material}~\ref{Supp:numer_sim}). Similarity $\mathbf{m}$=0 provides the best recognition performance \textbf{89.567} in terms of the AVG, then $\mathbf{m}$=0.1 has the AVG of 89.368, with $\mathbf{m}$=-0.1 obtains 87.708. It can be concluded the appropriate degree of similarity for generating discriminative samples is around 0 to 0.1, which is different from the similarity of CASIA-WebFace where the best recognition performance is obtained with the similarity of 0.7. This may be because the model for comparing similarity on CASIA-WebFace is pretrained on this dataset.


\textbf{Generation with Mixing m:}
We conduct the experiment of including mixed $\mathbf{m}$ when generating the dataset, as is shown in the top part of Tab~\ref{Tab:mix_m} and Tab~\ref{Tab:mix_m_largelevel}. Specifically, we opt for mixing the generation $\mathbf{m}$ from -0.1 to 0.1, and from 0 to 0.1. We use training m varying from [0,1], and the generation interval is 0.02. The results show that mixing m with 0 to 0.1 when generating the data will bring worse performance compared to single m=0. However, mixing m with -0.1 to 0.1 obtains a similar performance compared to m=0. Additionally, progressively mixing m with easy and semi-hard samples are provided in Tab~\ref{Tab:mix_m_largelevel}, as observed, with more easy samples included in the training dataset, the FR performance reduced more prominent. We keep the generation m to be 0 for later discussion.

\begin{table}[ht]
\begin{minipage}[b]{0.55\linewidth}
\resizebox{1.0\textwidth}{!}{
\begin{tabular}{@{}c|c|c|c|c@{}}
\toprule
Experiment &Training m                 & Generation m       & Interval  & AVG             \\ \midrule \midrule

\multirow{4}{*}{Mixing Generation m}&\multirow{4}{*}{{[}0,1{]}} & 0                  & 0.02        & \textbf{91.64}  \\
                    &
                           & {[}0,0.1{]}        & 0.02     & 91.11          \\
                           &
                           & {[}-0.1, 0.1{]}    & 0.02     & 91.61           \\ 
                           \midrule \midrule
\multirow{4}{*}{Mixing Training m}&{[}0,1{]}                  & \multirow{4}{*}{0} & 0.02     & 91.64           \\         &
0                          &                    & -        & 91.15 \\ 
&
{[}-1,1{]}                 &                    & 0.02     & 92.28 \\ 
&
{[}-1,1{]}                 &                    & 0.04     & \textbf{92.30} \\ 
&
{[}-1,1{]}                 &                    & 0.06     & 92.09 \\ \midrule 
\end{tabular}
}
\caption{Abaltion studies for mixing m in training and generation stage. The generation $\mathbf{m}$ is \textbf{mixed} with close similarity of \textbf{semi-hard} samples.}
\label{Tab:mix_m}
\end{minipage}
\hspace{5mm}
\begin{minipage}[b]{0.4\linewidth}

\resizebox{1.0\columnwidth}{!}{
\begin{tabular}{@{}c|cccc|c@{}}
\toprule
\multirow{2}{*}{Training m} & \multicolumn{4}{c}{Generation m}                                                                               & \multirow{2}{*}{AVG} \\ 
                            & 0                         & 0.5                       & 0.9                       & 1                         &                      \\ \midrule \midrule 
\multirow{4}{*}{{[}0,1{]}}  & \checkmark &                           &                           &                           & 91.64                \\
                            & \checkmark & \checkmark &                           &                           & 90.36                \\
                            & \checkmark & \checkmark & \checkmark &                           & 90.12                \\
                            & \checkmark &                           &                           & \checkmark & 89.57   \\ \midrule            
\end{tabular}}
\caption{Abaltion studies for \textbf{mixing} $\mathbf{m}$ in generation stage with \textbf{easy and semi-hard samples}}
\label{Tab:mix_m_largelevel}
\end{minipage}  

\end{table}

\textbf{Training with Various m:}
\label{aba:train_m}
The choice of various levels of $\mathbf{m}$ during the training stage are ablated in the bottom part of Table~\ref{Tab:mix_m} where 3 settings are considered when training CemiFace:(a) single $\mathbf{m}$ with similarity of 0; (b) multiple discrete $\mathbf{m}$ ranging from 0 to 1 with 50 steps; (c) similar to (b) but with a range of [-1,1] and interval 0.04. Then we synthesize the data with $\mathbf{m}$=0. As observed, setting (c) yields the best performance, indicating that with a broad range of similarity across -1 to 1, covering all the available probabilities, the CemiFace model can generalize well when adapted for generating highly discriminative samples. We also include experiments of changing the interval for (c) setting from 0.02 to 0.06 at the bottom of Tab~\ref{Tab:mix_m}, the result suggests that our approach is robust to the discrete interval but sensitive to the range of training $\mathbf{m}$. We do not consider continuous similarity as the trained model collapses to generate the same image when given different similarities $\mathbf{m}$.

\begin{table}[htbp]
\hspace{15mm}
\small
\begin{minipage}[t]{0.4\linewidth}
\small
\resizebox{1.0\columnwidth}{!}{
\begin{tabular}{@{}l|ll|l@{}}
\toprule
Method                  & Training Data           & Inquiry Data & AVG   \\ \midrule \midrule
\multirow{7}{*}{CemiFace}    & \multirow{3}{*}{CASIA}  & 1-shot Web  & \textbf{91.64} \\
                        &                         & DDPM        & 91.49 \\
                        &                         & 1-shot Flickr      & 88.97 \\
                        \cmidrule(l){2-4} 
                        & \multirow{3}{*}{Flickr} & 1-shot Web  & \textbf{90.25} \\
                        &                         & DDPM        & 90.19 \\
                        &                         & 1-shot Flickr      & 88.65 \\
                        \cmidrule(l){2-4} 
                        & \multirow{3}{*}{VGGFace2}                & 1-shot Web  & \textbf{92.20} \\
                        &                         & DDPM        & 92.01 \\
                        &                         & 1-shot Flickr      & 90.586 \\
                        \midrule \midrule
\multirow{2}{*}{DCFace} & \multirow{2}{*}{CASIA}  & 1-shot Web  & 89.8 \\
                        &                         & DDPM        & 90.18 \\ \cmidrule(l){1-4} 
\end{tabular}}
\caption{Impact of Training and Inquiry Data. We also include results of training on DCFace for comparison}
\label{Tab:data}
\end{minipage}
\hspace{15mm}
\begin{minipage}[t]{0.22\linewidth}

\resizebox{1.0\columnwidth}{!}{
\small
\begin{tabular}{@{}c|c|c@{}}
\toprule
Dataset                  & m    & AVG   \\ \midrule
\midrule
\multirow{3}{*}{WebFace} & -0.1 & 91.27 \\
                         & 0    & \textbf{91.64} \\
                         & 0.1  & 90.89 \\ \midrule
\multirow{3}{*}{DigiFace}                 & -0.1 & 89.96 \\
                         & 0    & \textbf{90.67} \\
                         & 0.1  & 90.38 \\ \midrule
\multirow{3}{*}{DDPM}    & -0.1 & 91.36 \\
                         & 0    & \textbf{91.47} \\
                         & 0.1  & 90.96 \\ \midrule 
\end{tabular}}

\caption{Accuracy of the optimal $\mathbf{m}$ on different inquiry sets.}
\label{Tab:m_crossdataset}
\end{minipage}  
\end{table}

\vspace{-8mm}
\subsubsection{Ablation Study for Training and Inquiry Data}\label{Sec:training_data}

\textbf{Impact of Training Data:} Since our method does not require paired images for training the diffusion model, the limitation of using unlabelled data is alleviated. Consequently, we conduct experiments to see the impact of different training data. Specifically, we employ 3 datasets for training:(a) CASIA-WebFace as used in DCFace;  (b) A challenging in-the-wild dataset Flickr with 1.2M images collected by us from Flickr website; (c) VGGFace2~\cite{cao2018vggface2} which is a large-scale dataset containing 3.3M clean images. Training $\mathbf{m}$ is set to vary within the range of [-1,1] while generation $\mathbf{m}$ is kept as 0. We do not consider data from FFHQ~\cite{karras2019style} due to restrictions on being applied for face recognition.

We can see from Table~\ref{Tab:data} using VGGFace2 as the training set produces the best performance when training a model on it, indicating that training on a large-scale dataset will bring more advance in generating discriminative dataset. However, to conduct a fair comparison with previous methods, we adopt CASIA-WebFace for the following studies. Additionally, although Flickr contains much more challenging conditions such as blurred, cartoon, and occluded faces, it results in similar performance compared to DCFace~\cite{kim2023dcface}, which proves the effectiveness of our proposed CemiFace.



\textbf{Impact of the Inquiry Data:}
The choice of appropriate inquiry image $\mathbf{x}$ which can be referred to as an initial point, is essential because we regard the generated group from the given $\mathbf{x}$ to be an independent identity group. DCFace employs a pre-trained DDPM model~\cite{ho2020denoising} trained on FFHQ to generate synthetic facial images. The style bank is sampled from a real-world dataset, e.g. CASIA-WebFace~\cite{li2017learning}. Their process involves a combination of synthetic facial data and a real dataset. In contrast to DCFace, our method has fewer constraints when referring to the source data. The source data can be either synthetic or real, and we ablate the impact of using synthetic data and real data.

For taking synthetic data as the inquiry samples, we use the samples from DCface to conduct a fair comparison, noted as DDPM. As for adopting real-data, we consider two options: (a) 1-shot data randomly sampled from WebFace-4m~\cite{zhu2021webface260m} which provides a clean dataset. (b) 1-shot Flickr, a challenging dataset filtered from the one collected in Sec.~\ref{Sec:training_data}, with fewer licence restriction. If inquiry images with high similarity, they result in overlapped groups of synthetic images in hypersphere space. Therefore,  we follow DCface to filter out samples with a similarity higher than 0.3. We ablate the choice of the inquiry data source in Table~\ref{Tab:data}, observing from changing the inquiry data, using 1-shot data of WebFace4M performs slightly better for our CemiFace. However, applying 1-shot WebFace4M to DCFace leads to a performance drop, as there are constraints for DCFace training and generation, e.g. frontal face and no glasses. Then using the challenging 1-shot Flickr as inquiry data brings worse results. This indicates that clean and real inquiry images are beneficial to generate discriminative datasets. Additionally, appropriate $\mathbf{m}$ for each inquiry dataset with 0.5M volume is also around 0 which can be observed in Tab~\ref{Tab:m_crossdataset}.

\begin{table}[h]
\centering
\resizebox{1.0\textwidth}{!}{
\small
\begin{tabular}{@{}l|l|lllll|l|l@{}}
\toprule
Method                 & Data Volume            & LFW            & CFP-FP         & AgeDB          & CALFW          & CPLFW          & AVG              & GtR \\ \midrule \midrule
CASIA-WebFace (AdaFace)          & \multirow{2}{*}{0.49M} & 99.42          & 96.56          & 94.08          & 93.32          & 89.73          & 94.62                           & -               \\
CASIA-WebFace (CosFace)$\dagger$ &                        & 99.3           & 94.87          & 94.35          & 93.15          & 89.65          & 94.26                          & 0               \\ \midrule \midrule
SynFace                & \multirow{5}{*}{0.5M}  & 91.93          & 75.03          & 61.63          & 74.73          & 70.43          & 74.75                       & 19.51               \\
DigiFace               &                        & 95.4           & 87.40          & 76.97          & 78.62          & 78.87          & 83.45                       & 10.81               \\
IDiff-Face               &                        & 98.00           & 85.47          & 86.43          & 90.65          & 80.45          & 88.20                      & 6.06               \\

DCFace                 &                        & 98.55          & 85.33          & 89.70          & 91.60          & 82.62          & 89.56                        & 4.70               \\
DCFace$\dagger$                &                        & 98.33          & 87.7           & 90.01          & 91.61          & 83.26          & 90.18                          & 4.08           \\
CemiFace, ours                    &                        & \textbf{99.03}& \textbf{91.06}& \textbf{91.33}& \textbf{92.42}& \textbf{87.65} & \textbf{92.30}                & \textbf{1.96}  \\ \midrule \midrule
DCFace                 & \multirow{3}{*}{1.0M}  & 98.83          & 88.40          & 90.45          & 92.38          & 84.22          & 90.86                        & 3.40               \\
DCFace$\dagger$                 &       & 98.88 & 89.71 & 91.25 & 92.15&   85.2     & 91.44                & 2.82          \\
CemiFace, ours                    &                        & \textbf{99.18} & \textbf{92.75} & \textbf{91.97}  & \textbf{93.01} & \textbf{88.42} &  \textbf{93.07}                         & \textbf{1.19}  \\ \midrule \midrule
DigiFace               & \multirow{4}{*}{1.2M}  & 96.17          & 89.81          & 81.10          & 82.55          & 82.23          & 86.37                        & 7.89               \\
DCFace                 &                        & 98.58          & 88.61          & 90.97          & 92.82          & 85.07          & 91.21                        & 3.05               \\
DCFace$\dagger$                &                        & 99.05 & 89.8  & 91.73 & 92.7  & 86.05 & 91.87                         & 2.39           \\
CemiFace, ours                    &   &    \textbf{99.22}       &  \textbf{92.84}        &    \textbf{92.13}        &    \textbf{93.03}        &      \textbf{88.86}     &  \textbf{93.22}             &    \textbf{1.04}             \\ \midrule 

\end{tabular}
}
\caption{Comparison with the previous methods. AVG is the average accuracy of the 5 evaluation datasets. GtR is the results compared to CASIA-WebFace with CosFace. Methods with $\dagger$ are the results reproduced by our settings }
\label{Tab:sota}
\end{table}

\vspace{-6mm}
\subsection{Comparison with the State-of-Art methods}
\subsubsection{Quantitative Results:}

We compare our CemiFace with the previous methods to demonstrate its effectiveness. The models compared are SynFace~\cite{qiu2021synface}, DigiFace~\cite{bae2023digiface}, IDiff-Face~\cite{boutros2023idiff} and DCFace~\cite{kim2023dcface} in both 0.5M, 1M and 1.2M image volumes. The loss for training the synthetic dataset is CosFace. For the CemiFace training set, we choose CASIA-WebFace to have a fair comparison with DCFace, training $\mathbf{m}$ ranges from -1 to 1 with 50 discrete steps, and generation $\mathbf{m}$ is 0. The results are available in Table~\ref{Tab:sota}. In 0.5 M protocol, our method exceeds the previous state-of-art method DCface in terms of all the evaluation datasets where we achieve significant improvement on pose-sensitive dataset CFP-FP~ and CPLFW by 3.36 and 4.39 respectively. And in the average protocol, we get 92.30 while DCFace is 90.18. Our method still cannot exceed the model trained on the real dataset CASIA-WebFace, but we reduce the GAP-to-Real error from 4.08 to 1.96 ($\frac{4.08 - 1.96}{4.08}=51.96\%$ relative error) compared to DCFace.  When it refers to the 1.0M and 1.2M settings, a similar phenomenon can be observed, our method surpasses DCFace on all protocols which reduces the Gap-to-Real by half, i.e. 1.59 and 1.36. In general, CemiFace behaves well on all verification accuracy and improves pose-related performance by a large margin.

\subsubsection{Qualitative Results}\label{subsec:qual_res}
\begin{figure}[h]
  \centering
   \includegraphics[width=1.0\linewidth]{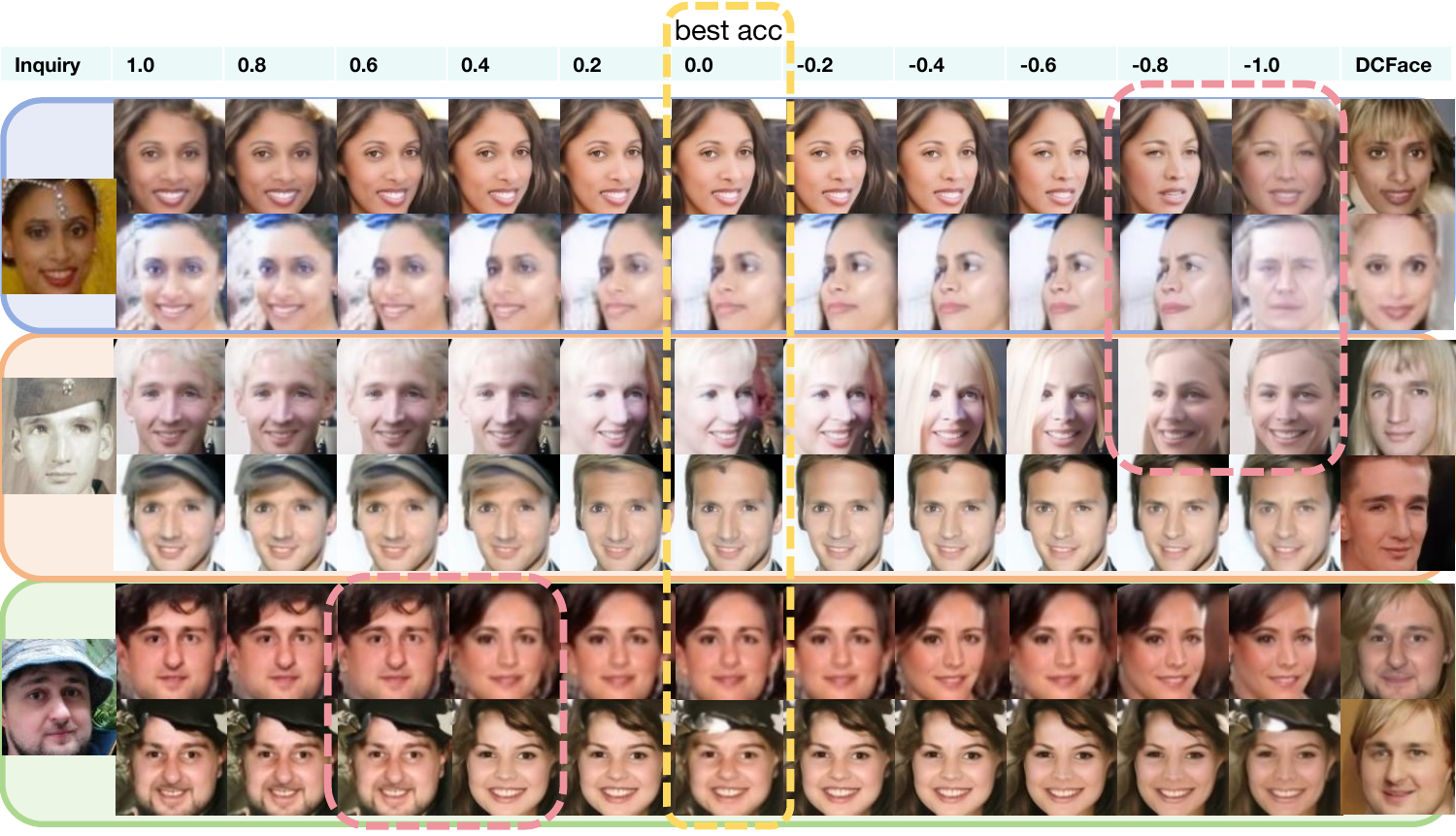}

    \caption{Sample Visualization under different similarity. From left to right are inquiry images, images with m from 1 to -1 and samples generated by DCFace. Different rows in each inquiry group represent the results produced by different noises. The first column are the inquiry images. The yellow dashed box includes samples where we obtain the best accuracy. Pink dashed boxes are samples that vary vastly. }
   
   \label{fig:visual_diff_sim_v2}
\end{figure}

We visualize the generated results to compare with DCFace in Figure~\ref{fig:visual_diff_sim_v2}. Specifically, samples with different $\mathbf{m}$ scaling between [-1,1] with interval 0.2 are presented. For each row, we opt for the same noise to illustrate the variations across different similarities. We observe that when $\mathbf{m}$ is set to 1, the identity of the generated sample is very close to the inquiry image. When $\mathbf{m}$ is 0.4, gender and age change can be observed from the last two rows. With $\mathbf{m}$ scaling far away from the inquiry image, pose changes can be noticed for the first 3 rows. Another interesting phenomenon appears when similarity is -1.0 where the generated samples change significantly. Additionally, when the noise changes, the generated images exhibit different styles, aligning with our hypothesis in Sec.~\ref{Sec:train_cemi}. Finally, with $\mathbf{m}$=0, the group looks extremely different to the inquiry image, but can deliver highly accurate face recognition performance.

\vspace{-3mm}

\vspace{-3mm}

\section{Conclusion}
This paper proposes a novel method to generate a discriminative dataset for training effective face recognition models with reduced privacy concerns. We investigate the factors contributing to the effective face recognition model training and re-formulate the challenge of generating discriminative samples as synthesizing center-based semi-hard samples. A similarity controlling factor condition is adopted for generating semi-hard samples. Models trained on the generated dataset with center-based semi-hard samples produce accurate face recognition performance over the previous methods. A notable advantage of CemiFace is its independence from a labelled dataset for training. However, the limitations of CemiFace include relying on the pretrained identity network's performance for conducting similarity comparisons, being sensitive to the quality of the inquiry image and privacy issues arising as the pretrained model derives from a dataset without user consent.


\paragraph{Acknowledgments:} This work was in part funded by UK Research and Innovation (UKRI) under the UK government’s Horizon Europe funding guarantee [grant number 10093336] and funded by the European Union [under EC Horizon Europe grant agreement number 101135800 (RAIDO)].

\bibliographystyle{unsrtnat}
\bibliography{nips}

\newpage
\appendix

This is the supplementary material for the paper \textbf{CemiFace: Center-based Semi-hard Synthetic Face Generation for Face Recognition}.

\section{Addition to: Implementation details}

\subsection{Diffusion Details}\label{Sec:runtime}
We follow most of the settings of DCFace~\cite{kim2023dcface}. Specifically, the model is trained on CASIA-WebFace~\cite{yi2014learning} with 10 epochs. The maximum time step $T$ for diffusion training is 1000. Then for generating the synthetic face recognition dataset, the time step for DDIM~\cite{song2020denoising} is 20. The optimizer opted for is AdamW~\cite{loshchilov2017decoupled}.
The batch size is 160 on 2 A100 GPUs. CemiFace training takes 8 hours, the generation also takes 8 hours. As a comparison, DCFace takes 10 hours for Training and 9 hours for Generation. Both DCFace and our method need around 6-7 hours to conduct FR training.

As for the diffusion UNet, we remove the identity feature in Residual Block, for more details of the Diffusion UNet please refer to DCFace~\cite{kim2023dcface}. 

\subsection{High Inter-class Variations and High Intra-class Variations}
\textbf{(1) High Inter-class Variations:} Each inquiry face image is selected to be highly independent on other inquiry images. Specifically, we follow DCFace to use a pre-trained FR model to keep samples with a threshold of lower 0.3.

\textbf{(2) High Intra-class Variations:} high intra-class variations are ensured by (a) changing the similarity condition $m$, as a small input similarity $m$ results in the generated semi-hard images belonging to the same identity having long distances to the identity center; and (b) the face images of the same identity generated by CemiFace are distributed in all directions from the identity center, which can be observed from \textbf{supplementary material} T-SNE Fig.~\ref{fig:T-sne}. This is guaranteed by randomly sampled Gaussian noises $\epsilon$ input to the diffusion model, which exhibits a large variation.  As a result, both properties would ensure the generated face images of the same identity are almost evenly distributed in a sphere that has a relatively large radius, and thus they would have high intra-class variations.

\subsection{Pseudo-code}\label{sec:pseudo}
The pseudo-code is provided below.

\begin{algorithm}\label{Alg:CemiFace_training}

	\renewcommand{\algorithmicrequire}{\textbf{Input:}}
	\renewcommand{\algorithmicensure}{\textbf{Output:}}
	\caption{The training pipeline of our CemiFace}  
	\label{LAFS:algorithm}
	\begin{algorithmic}[1]
            \STATE Initialization: Original Training Set $\mathbf{D_{o}}$, pretrained FR network $E_{\mathbf{id}}$, Diffusion Unet $\mathbf{\sigma_{\theta}}$, Maximum time step $T$, Maximum iteration $\tau$, iteration $n \leftarrow 0$, similarity $m\in [-1,1]$

            \REPEAT
            \STATE $n \leftarrow n + 1$
            \STATE Randomly sample a batch of facial images $x_{0}$ from $D_{o}$(also treated as inquiry data $d$),  noise images $\epsilon$ from normal distribution , similarity condition from range [-1,1], single time step $t$
            \STATE construct ID \& similarity condition $\mathbf{C_{att}}$ using Eq.~\ref{eq:attention}.
            \STATE add noise $x_{t} \leftarrow 
 \mathrm{use\ Eq.~\ref{eq:add_nosie}, given}\  x_{0} \& t$
            \STATE output estimated noise $\mathbf{\epsilon'}=\mathbf{\sigma_{\theta}}(\mathbf{x}_{t},t,\mathbf{C_{att}})$ 
            \STATE Update $\mathbf{\sigma_{\theta^{n+1}}} \leftarrow \mathbf{\sigma_{\theta^{n}}}- \nabla_{\mathbf{\sigma_{\theta}^{n}}}$ Eq.~\ref{eq:total_loss}

            \UNTIL converges or $n=\tau$
            
            \ENSURE output model $\mathbf{\sigma_{\theta}}$
 
	\end{algorithmic}  
\end{algorithm}

\begin{algorithm}

	\renewcommand{\algorithmicrequire}{\textbf{Input:}}
	\renewcommand{\algorithmicensure}{\textbf{Output:}}
	\caption{The pipeline of CemiFace-based face dataset generation}  
	\label{LAFS:algorithm2}
	\begin{algorithmic}[1]
            \STATE Initialization: Inquiry Data $D_{I}$, pre trained Diffusion Unet $\mathbf{\sigma_{\theta}}$, Maximum time step $T$, fixed similarity $m$, Maximum Number of samples in each identity $K$
            
            \STATE $n=0$ is the identity index, $k=0$ is the sample index
            \REPEAT
            \STATE $n \leftarrow n + 1$, $k=0$

            \STATE  Sample a batch of inquiry data $d$, construct the ID \& similarity condition $\mathbf{C_{att}}$ using Eq.~\ref{eq:attention}
            \REPEAT
            \STATE $k \leftarrow k+1$, $t=T$
            \STATE Generate noise image $x_{t}$ from normal distribution $N(0,I)$
            
            \REPEAT
            
            \STATE output estimated noise $\mathbf{\epsilon'}=\mathbf{\sigma_{\theta}}(\mathbf{x}_{t},t,\mathbf{C_{att}})$ 
            \STATE denoise the image using following DDIM~\cite{song2020denoising}
            $x_{t-1}\leftarrow \mathrm{denoise}(x_{t},\mathbf{\epsilon'})$
            \STATE $t \leftarrow t - 1$
            \UNTIL t=0
            \STATE assign $x_{0}$ the same label $y_{d}=n$ of the inquiry data, $[x_{0},y_{d}]$
            \UNTIL k=K
            \UNTIL $n=\mathrm{len}(D_{i})$
            
            \ENSURE output the generated dataset

	\end{algorithmic}  
\end{algorithm}

\subsection{Dataset statistics}\label{Sec:dataset_sta}

We have also calculated the number of face images belonging to different similarity groups for CemiFace and DCFace in the Tab~\ref{Tab:Num_similaritygroup}, indicating that our CemiFace tends to generate images showing lower similarities to their identity centers (i.e. all samples are semi-hard), while DCFace containing more easy samples.

\begin{table}[h]
\centering
\resizebox{1.0\columnwidth}{!}{
\begin{tabular}{@{}l|ll|llllll@{}}
\toprule
\multirow{2}{*}{Method} & \multirow{2}{*}{avg sim} & \multirow{2}{*}{std} & \multicolumn{6}{l}{Number of identites} \\ \cmidrule(l){4-9} 
                        &                              &                      & 0-0.1  & 0.1-0.2  & 0.2-0.3  & 0.3-0.4  & 0.4-0.5  & above 0.5 \\ \midrule
DCFace                  & 36.24                        & 9.14                 & 14     & 231      & 2059     & 5899     & 1788     & 9         \\
CemiFace                & 28.54                        & 7.76                 & 196    & 1043     & 3281     & 4539     & 930      & 7         \\ \bottomrule
\end{tabular}}
\caption{The statistics of the average similarity of each group. \textbf{avg sim} and \textbf{std} is the average/std similarity to the inquiry images of the whole dataset. \textbf{0-0.1} means the number of identities has a similarity of 0-0.1. CemiFace is distributed farther away from the inquiry center with less variation than DCFace.}
\label{Tab:Num_similaritygroup}
\end{table}

\section{Further Experiments}\label{Sec:further_dis}

\subsection{Impact of Identity Center and Random Center}
The performance of CemiFace is highly affected by the characteristics of the inquiry samples. Herein we examine how the model behaves when subjected to numerical identity conditions. Two kinds of centers are considered:(a) identity centers derived from the CASIA-WebFace dataset, and (b) random centers with a similarity range of [-0.1, 0.2] to (a). By observing from the Table~\ref{tab:random_class_center}, with random center the model results in invalid results; On the other hand, when utilizing identity centers, the model performs optimally when the similarity controlling condition $\mathbf{m}$ is set to 0 which aligns our previous finding. However, it is noteworthy that with identity center the performance is worse than the dataset inquired by 1-shot WebFace, exhibiting similar results to DCFace. 

\begin{table}[h]
\centering
\small
\resizebox{0.6\textwidth}{!}{
\begin{tabular}{@{}l|l|lllll|l@{}}

\toprule
Inquiry source                 & sim & LFW   & CFP-FP & AgeDB & CALFW & CPLFW & AVG   \\ \midrule \midrule
Random  Center               & 1.0 & \multicolumn{5}{c}{Not converge}       &       \\ \midrule
\multirow{6}{*}{Identity Center} & 1.0 & 96.80  & 71.81    & 86.13    & 89.52     &  71.72  &   83.20    \\
                       & 0.7 & 97.22 & 75.03  & 86.90 & 89.93 & 74.47 & 84.71 \\
                       & 0.5 & 97.50 & 78.96  & 87.12 & 90.38 & 77.62 & 86.32 \\
                       & 0.2 & 98.17 & 86.29  & 89.07 & 91.40 & 83.03 & 89.59 \\
                       & 0.1 & \textbf{98.25} & 87.30  & \textbf{89.98} & 91.35 & 83.23 & 90.02 \\
                       & 0.0   & 98.23 & \textbf{87.49}  & 89.53 & \textbf{91.47} & \textbf{83.73} & \textbf{90.09} \\ \midrule \midrule
1-shot DigiFace             & 0.0   & 98.28 & 90.04  & 89.68 & 91.23 & 84.12 & 90.67 \\
1-shot WebFace             & 0.0   & 99.03 & 91.06  & 91.33 & 92.42 & 87.65 & \textbf{92.30} \\
DCFace                 & -   & 98.33 & 87.7   & 90.01 & 91.61 & 83.26 & 90.18 \\ \bottomrule
\end{tabular}
}
\caption{Comparison of different inquiry centers. The results of DCFace run by our setting are copied for reference.}
\label{tab:random_class_center}
\end{table}

\begin{figure}[h]
  \centering
   \includegraphics[width=0.85\linewidth]{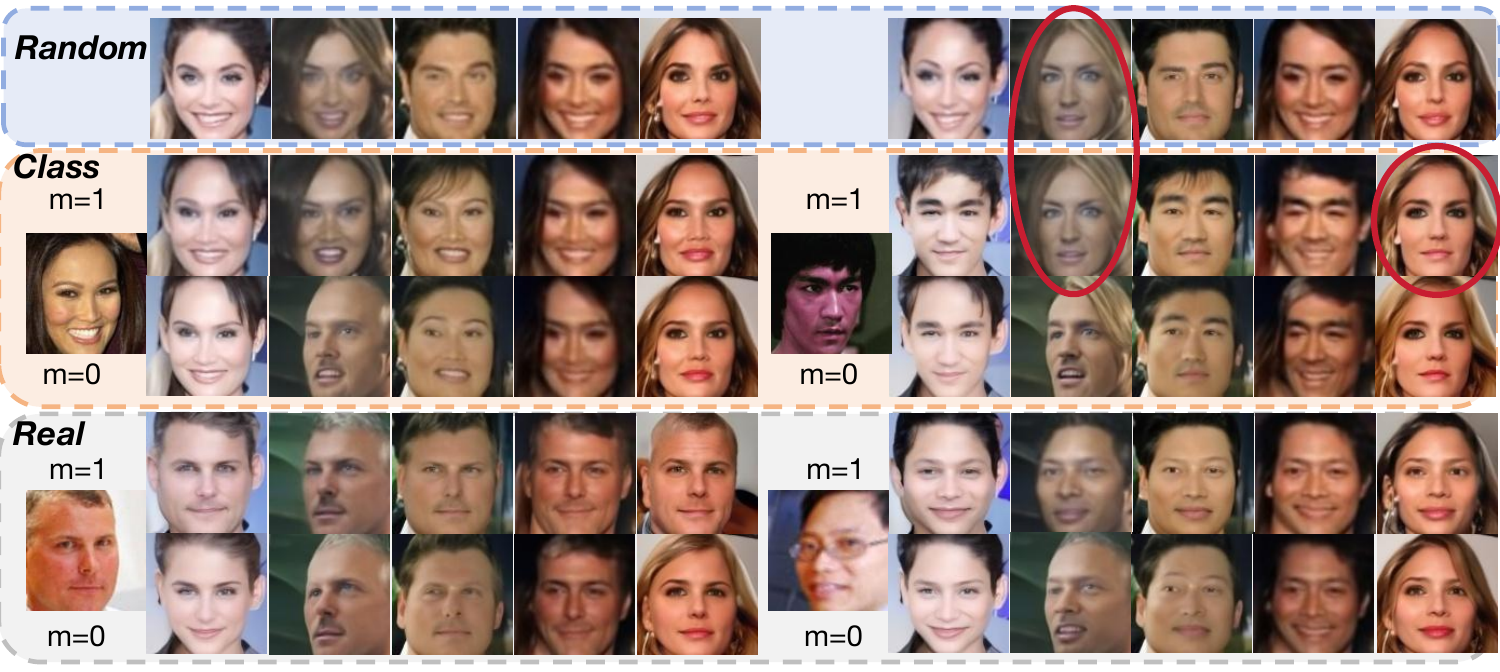}

    \caption{Comparison of different inquiry center. From top to bottom are images inquired by \textit{Random Center}, CASIA \textit{Identity Center} and 1-shot \textit{Real} images. For \textit{Identity Center} and 1-shot \textit{Real} images, images similarity of 1 and 0 are shown. Different columns represent given different noise. Two examples are shown for each case. The inquiry images in the identity center are selected from the dataset. The red circles contain samples that look extreme different from the inquiry center.}
   
   \label{fig:center_compare}
\end{figure}

To provide deeper insights into this phenomenon, we visualize the samples generated by different inquiry centers in Figure~\ref{fig:center_compare}. Notably, with $\mathbf{m}$=1 the random center produces images with different identities which can simply be concluded by human observation. Conversely with the identity center, given a similarity of 1.0, the generated samples appear highly similar, except for the samples circled in red. Further investigation reveals that the number of images in that subject comprises approximately 16 images while the left subject provides approximately 50 images. Intuitively, A model trained on this dataset will focus more on the subjects with a large number of images which explains the suboptimal results obtained by identity center. 

\begin{figure}[h]
  \centering
   \includegraphics[width=0.5\linewidth]{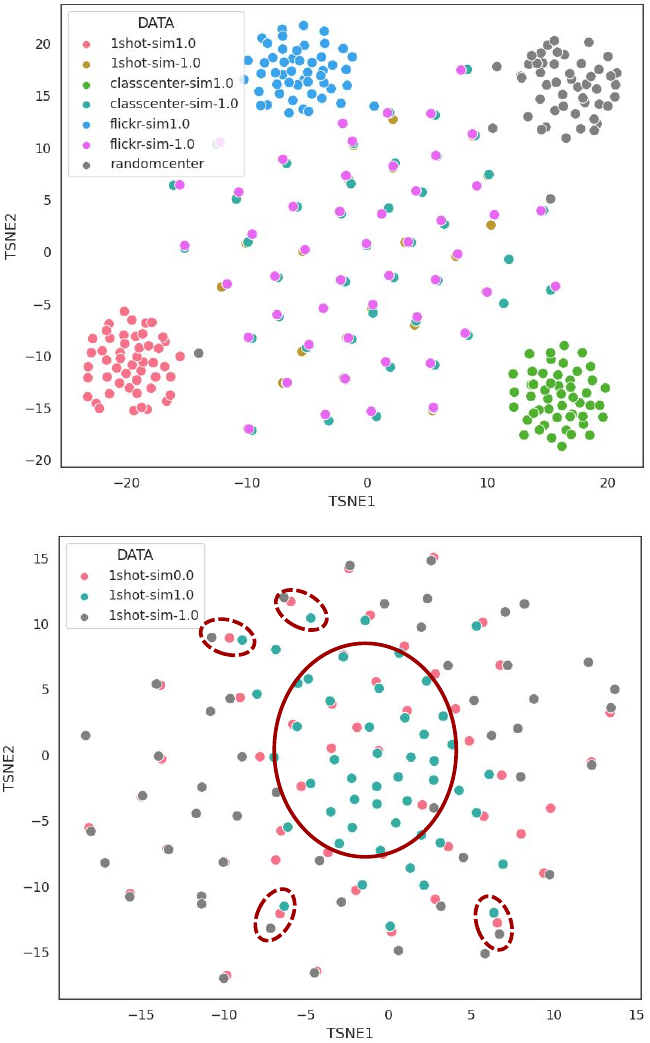}

    \caption{ T-SNE visualization. The bottom figure is the T-SNE generated by 1-shot data with similarity of 1.0, 0.0 and -1.0 respectively.  The upper figure is different inquiry centers with two similarities 1.0 and -1.0, the random center is also given. Red circles are samples worth noticing, with their order being green, red, and grey, positioned from center to outside }
   
   \label{fig:T-sne}
\end{figure}

We further visualize the T-SNE of the feature embedding in Figure~\ref{fig:T-sne}. As shown in the upper figure, with higher similarity, the samples tend to cluster in the central region. Subsequently, by inspecting the bottom figure, it becomes apparent that with a similarity of 1.0, each subject is located in a different specific area. Consequently, a similarity of -1.0 results in each image being positioned close to other subjects in the middle area.

\subsection{Addition to the Inquiry Data: Image Quality}
\begin{figure}[htbp]
  \centering
   \includegraphics[width=0.5\linewidth]{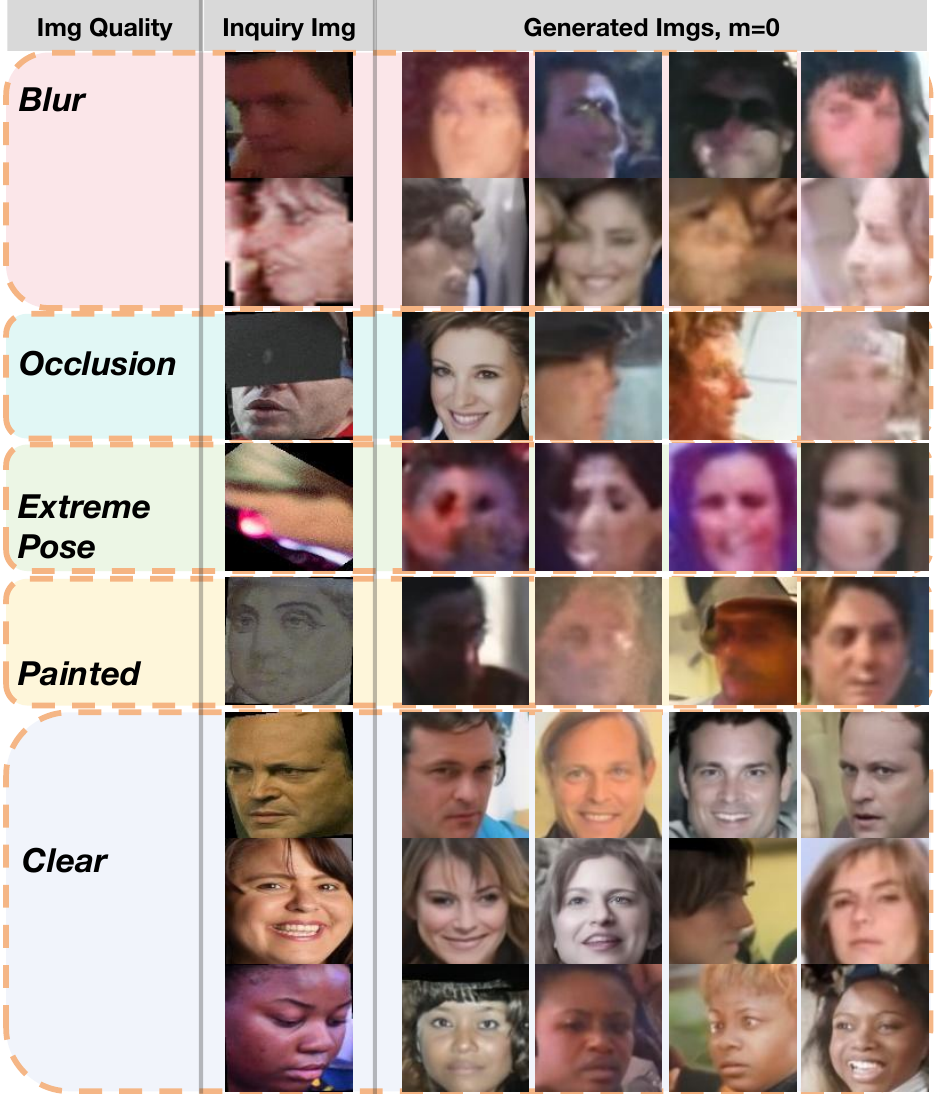}

    \caption{Examples of samples under challenging conditions, including Blur, Occlusion, Extreme Pose, and Painted conditions, are presented. Samples generated by clear images are appended for better comparison.}
   
   \label{fig:image_quality}
\end{figure}

The above discussion validates how CemiFace is affected by different centers in the aspect of numerical results. For a better understanding of the negative impact brought by challenge inquiry data such as 1-shot Flickr, we visualize the images generated from different image quality in Figure~\ref{fig:image_quality}. Specifically, we present inquiry images subjected to \textit{blur}, \textit{occlusion}, \textit{extreme pose}, \textit{painted} and \textit{clear} conditions, with a similarity controlling condition $\mathbf{m}$ set to 0. By comparing the last block with the rest of the blocks, one can conclude that extreme image quality fails to generate clean images. In conclusion, unblurred, non-occluded, appropriately posed, and real-world data are essential for our model to generate a highly clean synthetic face recognition dataset.

\subsection{Further Ablation Studies}
\subsubsection{Impact of Different Pretrained loss} 

As DCFace hasn't released its AdaFace-based SFR training code and details, we were not able to reproduce it for our model training. Thus, in Tab~\ref{Tab:sota} fairly compare ours with DCFace by adopting the same pre-trained AdaFace model to train our diffusion generator, and then employing the same CosFace loss for both ours and DCFace's SFR models training. Results show that our CemiFace still outperformed the SOTA DCFace. Additionally, we provide results achieved by using pre-trained model trained by CosFace. Specifically, we apply a model pre-trained by CosFace to train both our generator, and employ the same CosFace loss for their SFR models' training. The experiment shows that the model pretrained from CosFace performs better than that of AdaFace.

\begin{table}[h]
\centering
\begin{tabular}{@{}l|l|l|l@{}}
\toprule
Method        & Pretrained FR            & SFR loss                 & AVG            \\ \midrule \midrule
CASIA-WebFace & -                        & AdaFace                  & 94.62          \\
CASIA-WebFace & -                        & CosFace                  & 94.26          \\ \midrule
CemiFace      &   AdaFace                       &  CosFace                        & 92.30          \\ \midrule
CemiFace      &    CosFace                      &  CosFace                        & \textbf{92.60} \\ \cmidrule(r){1-4} 
\end{tabular}
\end{table}

\subsection{Upper/Lower Bound of Different Similarity Group in CASIA-WebFace dataset}\label{Sec:sim_range}
The range of each similarity group in the Section~\ref{ab:m} is given in the following Table~\ref{Tab:similarity_range}

\begin{table}[h]
\centering
\resizebox{0.7\columnwidth}{!}{
\begin{tabular}{@{}llll@{}}
\toprule
Avg Sim & average largest sim & average lowest sim & AVG   \\ \midrule
0.85    & 0.887               & 0.831              & 89.48 \\
0.81    & 0.831               & 0.794              & 91.01 \\
0.76    & 0.794               & 0.747              & 91.78 \\
0.70    & 0.747               & 0.676              & 91.55 \\
0.53    & 0.767               & 0.277              & 82.36 \\ \bottomrule
\end{tabular}}
\caption{ \textbf{Average largest sim} represents the mean value of the largest similarity values appeared in every identity; and \textbf{Average lowest sim} represents the mean value of the lowest similarity values appeared in every identity}
\label{Tab:similarity_range}
\end{table}

\subsubsection{Impact of Different Training Backbone} Following previous works(DCFace~\cite{kim2023dcface}, DigiFace~\cite{bae2023digiface}, SynFace~\cite{qiu2021synface}), we use the IResnet-SE-50 modified by ArcFace~\cite{deng2019arcface} as the default backbone. Additionally, we provide the results achieved by IResnet-18(R18), IResnet-SE-50(R50) and IResnet-SE-100(R100) in table~\ref{fig:backbone} for reference.
\begin{table}[h]
\centering
\small
\begin{tabular}{@{}llll@{}}
\toprule
Backbone & R18   & R50   & R100  \\ \midrule
AVG      & 90.75 & 91.64 & 91.82 \\ \bottomrule
\end{tabular}
\caption{Impact of different training backbone}
\label{fig:backbone}
\end{table}

\subsubsection{Numercial Results for Different $\mathbf{m}$}\label{Supp:numer_sim}
\begin{table}[h]
\centering
\tiny
\begin{tabular}{@{}c|ccccc|c@{}}
\toprule
Sim  & LFW   & CFP-FP & AgeDB-30 & CALFW & CPLFW & AVG    \\ \midrule \midrule
1    & 97    & 72.94  & 86.98    & 89.85 & 73.86 & 84.126 \\
0.9  & 97.38 & 73.81  & 86.88    & 90.13 & 74.82 & 84.604 \\
0.8  & 85.75 & 62.42  & 67.8     & 81.85 & 58.43 & 71.25  \\
0.7  & 97.2  & 75.5   & 86.75    & 90.15 & 75.95 & 85.11  \\
0.6  & 97.52 & 78.91  & 87.25    & 90.84 & 75.39 & 85.982 \\
0.5  & 97.85 & 80.55  & 87.93    & 90.9  & 79.35 & 87.316 \\
0.4  & 97.88 & 80.39  & 88.01    & 90.89 & 79.55 & 87.344 \\
0.3  & 97.98 & 80.19  & 88.15    & 90.72 & 79.73 & 87.354 \\
0.2  & 98.02 & 84.21  & 88.6     & 91.03 & 81.99 & 88.77  \\
0.1    & 98.2  & 86.29  & 88.25    & 91.25 & 82.85 & 89.368 \\ \midrule
0  & 98.1  & 86.6   & 88.9     & 91.15 & 83.08 & \textbf{89.567} \\ \midrule
-0.1 & 97.65 & 84.9   & 86.42    & 89.47 & 80.1  & 87.708 \\
-0.2 & 93.15 & 80.83  & 81.33    & 85.92 & 74.68 & 83.182 \\
-0.3 & 92.77 & 74.13  & 78.15    & 81.58 & 69.72 & 79.27  \\
-0.4 & 89.11 & 71.78  & 70.13    & 77.78 & 65.17 & 74.794 \\
-0.5 & 85.18 & 65.16  & 63.42    & 69.58 & 63.68 & 69.404 \\
-0.6 & 84.23 & 64.63  & 63.05    & 69.13 & 62.86 & 68.78  \\
-0.7 & 83.65 & 63.98  & 62.53    & 68.78 & 61.26 & 68.04  \\
-0.8 & 82.1  & 62.51  & 61.85    & 67.53 & 60.7  & 66.938 \\
-0.9 & 84.23 & 62.38  & 65.13    & 73.85 & 60.08 & 69.134 \\
-1   & 85.75 & 62.42  & 67.8     & 81.85 & 58.43 & 71.25  \\ \bottomrule
\end{tabular}
\caption{Numercial results for the impact of different similarities}
\label{Tab:num_m}
\end{table}
Here we provide the numerical results for the impact of different similarity levels in Tab~\ref{Tab:num_m}, $\mathbf{m}=0$ provide the best performance.




\subsubsection{FID Image Quality}
We use Fréchet Inception Distance(FID) which measures the distribution similarity of the given two datasets.
Specifically, in Tab~\ref{Tab:Fid}, FID is reported by comparing randomly selected 10k samples with randomly selected CASIA. Need to note that our method doesn’t intend to generate images similar to the distribution of CASIA-WebFace, but to construct a discriminative dataset that is conducive to providing highly accurate FR performance

\begin{table}[h]
\centering
\small
\begin{tabular}{@{}c|ccc@{}}
\toprule
Method & Ours  & DCFace~\cite{kim2023dcface} & DigiFace~\cite{bae2023digiface} \\ \midrule
FID    & 18.72 & 15.82  & 65.39    \\ \bottomrule
\end{tabular}
\caption{Fid score to the real dataset CASIA-WebFace.}
\label{Tab:Fid}
\end{table}

\subsubsection{Euclidean Distance}
As shown in Tab~\ref{Tab:euc_inter} using Euclidean distance leads to worse performance than cosine similarity, which might be due to the FR training loss (CosFace~\cite{wang2018cosface}) being carried on cosine similarity.
\begin{table}[h]
\centering
\small
\begin{tabular}{@{}c|c|c@{}}
\toprule
Base  & Euclidean & Interval 0.06 \\ \midrule
\textbf{91.64} & 90.95     & 91.43              \\ \bottomrule
\end{tabular}
\caption{Difference between Euclidean and larger similarity interval}
\label{Tab:euc_inter}
\end{table}

\subsubsection{Impact of $\lambda$}
We present the results using different $\lambda$ in the left part of the Tab~\ref{tab:lamda}. Performance is sensitive to $\lambda$, and large $\lambda$ results in performance degradation.

\begin{table}[h]
\centering
\begin{tabular}{@{}c|cccc@{}}
\toprule
$\lambda$ & 0.01 & 0.05(default) & 0.1 & 0.5 \\ \midrule
AVG    &   \textbf{91.72}   & 91.64         & 91.29    &  90.77   \\ \bottomrule
\end{tabular}
\caption{Impact of different $\lambda$}
\label{tab:lamda}
\end{table}

\section{Privacy Concerns}\label{Sec:pri_con}
In this section, we are going to discuss the privacy issues that lie in developing synthetic face generation for face recognition. The primary aim of synthetic face recognition is to mitigate concerns associated with privacy. Large-scale face recognition data are usually collected from web scrappers by searching name lists (usually celebrities), without obtaining user consent. Consequently, some of the large-scale datasets~\cite{cao2018vggface2,guo2016ms} are abandoned by their collector to avoid Legal Risk. In addition, IDiff-Face~\cite{boutros2023idiff} mentions European Union (EU) has come up with the General Data Protection Regulation (GDPR)~\cite{regulation2016regulation} to regulate the application of facial data, making it harder to use face recognition data. 

\begin{figure}[htbp]
  \centering
   \includegraphics[width=0.5\linewidth]{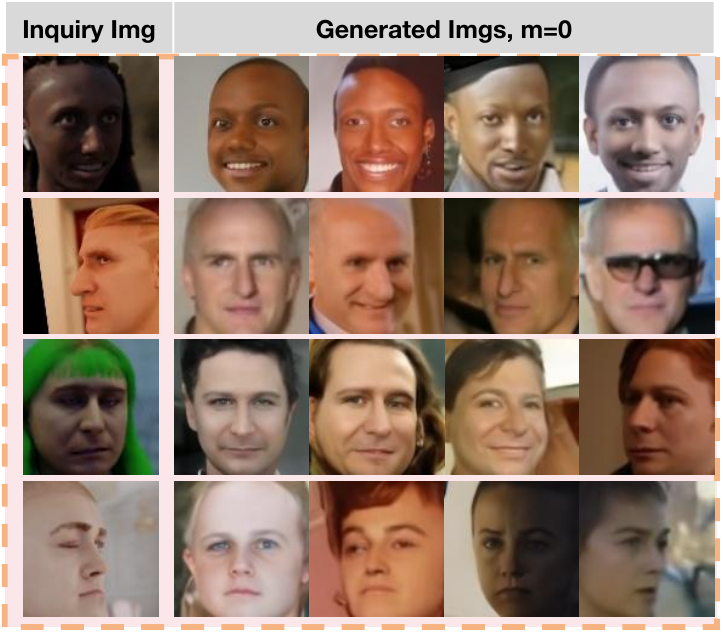}

    \caption{visualization of samples inquired by 1-shot DigiFace. Different rows are results inquired by different images. Different columns are randomly selected generated samples.}
   
   \label{fig:digiface}
\end{figure}

We notice that DCFace~\cite{kim2023dcface} incorporates a labelled dataset for training style transferring solution, and when they generate the new dataset, they use samples provided by DDPM~\cite{ho2020denoising} trained on FFHQ~\cite{karras2019style}. However, a noteworthy concern arises as the FFHQ dataset, whose derivative model is used as pretrained model in DCFace for sample generation, explicitly bans its application in face recognition. Consequently, we are not sure whether the model and synthetic face images based on FFHQ are allowed to be used. We try to avoid privacy concerns from the aspect of collecting Flickr which contains diverse licenses with reduced privacy problems. Another potential solution to avoid privacy concerns is to use samples like Digiface~\cite{bae2023digiface} which is rendered by 3DMM. However, DigiFace is only allowed to be adopted for non-commercial applications, but one can render images from 3DMM following the DigiFace pipeline for commercial purposes. We append the result inquired by 1-shot DigiFace in the bottom part of Table~\ref{tab:random_class_center} for reference and example images generated by 1-shot DigiFace are shown in Figure~\ref{fig:digiface}. Results reveal that 1-shot DigiFace still can not surpass 1-shot WebFace but still behave better than DCFace. Finally, although 1-shot Digiface samples sometimes don't appear to be like real humans, the generated samples exhibit similar patterns to real-world images from human observation.

Our method CemiFace offers the advantage of not requiring labels during the training phase compared to DCFace. Nonetheless, both our method and DCFace adopt a pre-trained face recognition model which may counter legal issues. we hope further researchers bring steps forward to avoid using this kind of pre-trained face recognition model to alleviate legal concerns in this domain.

\section{Discussion}

\subsection{Why Semi-hard samples work}
We assume the benefits of the semi-hard training face images could be attributed to: 
\begin{itemize}
    \item easy training samples are typically images where the face is clear, well-lit, and faces the camera directly, and thus training on such easy samples would not allow the trained FR models to be able to generalize for face images with large pose/age/expression variations and different lighting conditions/backgrounds that are frequently happened in real-world applications. AdaFace~\cite{kim2022adaface} also mentioned that easy samples could be beneficial to early-stage training, while hard sample mining is needed for achieving generalized and effective FR models; \\
    \item Hard samples normally contain noise data. Specifically, FaceNet~\cite{schroff2015facenet} demonstrates that the hardest sample mining using a large batch size leads to hard convergence and produces inferior performance. This is because training with very hard samples may not allow FR models to learn effective features but focus on cues apart from facial identities; 
    \item Semi-hard samples generated by CemiFace mostly contain large posed faces but fewer face-unrelated noises. We also evaluate the training epochs needed to reach the highest AVG performance for easy samples ($m=0.7$), semi-hard samples($m=0$) and extreme hard samples ($m=-0.5$). Easy samples take 10 epochs to reach the best AVG and 20 epochs to produce the training loss of 0; Semi-hard samples take much longer (38 epochs) to provide the highest AVG while the final training loss is around 3; and FR models training on extreme hard samples could not converge.
\end{itemize}
The actual similarity to the inquiry center indicates that our CemiFace tends to generate images showing lower similarities to their identity centers (i.e. all samples are semi-hard), while DCFace contains more easy samples.

\subsection{Different diffusion Loss}
As there are some other variation diffusion losses such as Improved-DDPM~\cite{nichol2021improved} which has been applied in Diffusion Transformer ( DIT)~\cite{peebles2023scalable}, Variational Diffusion Models (VDM)\cite{kingma2021variational}. We follow the previous SOTA SFR studies (DCFace~\cite{kim2023dcface} and IDiffFace~\cite{boutros2023idiff}) to choose the same generic MSE diffusion loss~\cite{ho2020denoising,song2020denoising} as our base model, ensuring the reproducibility of our approach and its fair comparison with DCFace~\cite{kim2023dcface} and IDiffFace~\cite{boutros2023idiff}.

\subsection{Difference between Dataset Distillation}
Dataset distillation methods~\cite{yin2023squeeze,wang2018dataset,loo2022efficient} are widely adopted to create a dataset that can produce high performance when training a model on it. SRe2L~\cite{yin2023squeeze} is a recent state-of-the-art method for dataset distillation which trains the noise image through a pretrained backbone. Their main process contains a forward process to get the classification label of the trainable noise inquiry image and train the noise inquiry image to produce a specific class prediction with BN alignment. The distinctions between our method with theirs are:

\begin{itemize}
    \item \textbf{Embedding vs Classification Layer}: We aim to explore the feature embedding of the backbone, not the classification layer.

    \item \textbf{Consideration of Image Similarity}: Our method explores the similarity of the given inquiry image, which is not considered in recent dataset distillation methods.

    \item \textbf{Pattern Distillation}: Their approach focuses on distilling data from existing classes, while our CemiFace distils patterns from the pretrained face recognition model. This learned pattern can be applied to unseen subjects, as we utilize independent data that was not part of the pretrained model's training dataset. 

    \item \textbf{Extra Model}: We incorporate a diffusion model to introduce parameters for producing high-quality images.
\end{itemize}





\subsection{Relationship to ID3PM}
Recent work, i.e. ID3PM~\cite{kansy2023controllable} proposes to invert the Black-Box model of face recognition to generate a similar image to the inquiry image. However, our method differs from theirs in several aspects:
\begin{itemize}
    \item \textbf{Purpose}: Their objective is to invert the black-box model without full access, whereas we aim to generate a discriminative dataset.
    \item \textbf{Image Similarity}: They require the generated image to be like the original image, while our goal is to ensure the generated images encompass diverse styles.
    \item \textbf{Evaluation Approach}: They evaluate by replacing the data of the evaluation dataset, whereas our approach involves training a model on the generated dataset.

    \item \textbf{Theoretical Degradation}: When $\mathbf{m}$ is set to 1, our model theoretically degrades to their model.

    \item \textbf{Diffusion Model Structures}: We use different diffusion model structures to conduct experiments, specifically employing cross-attention and AdaGN~\cite{preechakul2022diffusion} for inserting conditions.
\end{itemize}






\newpage
\section*{NeurIPS Paper Checklist}



\begin{enumerate}

\item {\bf Claims}
    \item[] Question: Do the main claims made in the abstract and introduction accurately reflect the paper's contributions and scope?
    \item[] Answer: \answerYes{}
    \item[] Justification: Contributions of this paper are included in the abstract and introduction

\item {\bf Limitations}
    \item[] Question: Does the paper discuss the limitations of the work performed by the authors?
    \item[] Answer: \answerYes{}
    \item[] Justification: We have discussed the limitation in the conclusion section

\item {\bf Theory Assumptions and Proofs}
    \item[] Question: For each theoretical result, does the paper provide the full set of assumptions and a complete (and correct) proof?
    \item[] Answer: \answerNA{} 
    \item[] Justification: We only have experimental assumptions and they are proved in the main paper

    \item {\bf Experimental Result Reproducibility}
    \item[] Question: Does the paper fully disclose all the information needed to reproduce the main experimental results of the paper to the extent that it affects the main claims and/or conclusions of the paper (regardless of whether the code and data are provided or not)?
    \item[] Answer: \answerYes{}
    \item[] Justification: We will release the code and data upon acceptance. And we provide detailed information for reproducing.

\item {\bf Open access to data and code}
    \item[] Question: Does the paper provide open access to the data and code, with sufficient instructions to faithfully reproduce the main experimental results, as described in supplemental material?
    \item[] Answer: \answerNo{}
    \item[] Justification: We will release the code and data upon acceptance.

\item {\bf Experimental Setting/Details}
    \item[] Question: Does the paper specify all the training and test details (e.g., data splits, hyperparameters, how they were chosen, type of optimizer, etc.) necessary to understand the results?
    \item[] Answer: \answerYes{}
    \item[] Justification: All experimental details are provided

\item {\bf Experiment Statistical Significance}
    \item[] Question: Does the paper report error bars suitably and correctly defined or other appropriate information about the statistical significance of the experiments?
    \item[] Answer: \answerYes{}
    \item[] Justification: Our experimental results show that the proposed method exceeds previous works by a large margin. And we have run the experiments multiple times to confirm the effectiveness of the proposed method.

\item {\bf Experiments Compute Resources}
    \item[] Question: For each experiment, does the paper provide sufficient information on the computer resources (type of compute workers, memory, time of execution) needed to reproduce the experiments?
    \item[] Answer: \answerYes{}
    \item[] Justification: Computational cost is included in the Supplementary Material
    
\item {\bf Code Of Ethics}
    \item[] Question: Does the research conducted in the paper conform, in every respect, with the NeurIPS Code of Ethics \url{https://neurips.cc/public/EthicsGuidelines}?
    \item[] Answer: \answerYes{}
    \item[] Justification: We have made sure the anonymity

\item {\bf Broader Impacts}
    \item[] Question: Does the paper discuss both potential positive societal impacts and negative societal impacts of the work performed?
    \item[] Answer: \answerYes{}
    \item[] Justification: We have discussed the privacy issues brought by Face Recognition.
    
\item {\bf Safeguards}
    \item[] Question: Does the paper describe safeguards that have been put in place for responsible release of data or models that have a high risk for misuse (e.g., pretrained language models, image generators, or scraped datasets)?
    \item[] Answer:\answerYes{}
    \item[] Justification: We avoid using data that is banned from being applied to Face recognition and discussed the privacy issues.

\item {\bf Licenses for existing assets}
    \item[] Question: Are the creators or original owners of assets (e.g., code, data, models), used in the paper, properly credited and are the license and terms of use explicitly mentioned and properly respected?
    \item[] Answer:\answerYes{}
    \item[] Justification: Our models and code used in the paper are licensed

\item {\bf New Assets}
    \item[] Question: Are new assets introduced in the paper well documented and is the documentation provided alongside the assets?
    \item[] Answer: \answerNA{}
    \item[] Justification: No assets

\item {\bf Crowdsourcing and Research with Human Subjects}
    \item[] Question: For crowdsourcing experiments and research with human subjects, does the paper include the full text of instructions given to participants and screenshots, if applicable, as well as details about compensation (if any)? 
    \item[] Answer:\answerNA{}
    \item[] Justification: We do not collect data, but generate synthetic data

\item {\bf Institutional Review Board (IRB) Approvals or Equivalent for Research with Human Subjects}
    \item[] Question: Does the paper describe potential risks incurred by study participants, whether such risks were disclosed to the subjects, and whether Institutional Review Board (IRB) approvals (or an equivalent approval/review based on the requirements of your country or institution) were obtained?
    \item[] Answer: \answerNA{}
    \item[] Justification: We do not collect data, but generate synthetic data

\end{enumerate}

\end{document}